\documentclass[11pt]{article}

\usepackage[utf8]{inputenc}
\usepackage[T1]{fontenc}

\DeclareUnicodeCharacter{2014}{-}
\usepackage[margin=1in]{geometry}
\usepackage{graphicx}
\usepackage{booktabs}
\usepackage{multirow}
\usepackage{amsmath}
\usepackage{amssymb}
\usepackage{siunitx}
\usepackage{hyperref}
\usepackage{enumitem}
\usepackage[numbers]{natbib}
\usepackage{float}
\usepackage{subcaption}
\usepackage{fvextra}
\usepackage{listings}
\usepackage{microtype}

\hypersetup{
  colorlinks=true,
  linkcolor=blue,
  citecolor=blue,
  urlcolor=blue
}

\fvset{breaklines=true,breakanywhere=true}
\newcommand{\benchmarkname}{\textsc{VB}}
\newcommand{\benchlong}{Visibility Benchmark}

\newcommand{\maybeincludegraphics}[3][]{%
  \IfFileExists{#2}{%
    \includegraphics[#1,width=#3]{#2}%
  }{%
    \fbox{%
      \parbox[c][0.12\textheight][c]{#3}{%
        \centering\scriptsize
        Image placeholder\\
        \texttt{\detokenize{#2}}%
      }%
    }%
  }%
}

\title{\benchmarkname: \benchlong{} for Visibility and Perspective Reasoning in Images}
\author{
  Neil Tripathi\\
  New York University\\
  \texttt{nt2439@nyu.edu}
}
\date{March 2026}

\begin{document}
\maketitle

\begin{abstract}
We present \benchmarkname, a benchmark that tests whether vision-language models can determine what is and is not visible in a photograph---and abstain when a human viewer cannot reliably answer. Each item pairs a single photo with a short yes/no visibility claim; the model must output \texttt{VISIBLY\_TRUE}, \texttt{VISIBLY\_FALSE}, or \texttt{ABSTAIN}, together with a confidence score. Items are organized into 100 families using a \(2 \times 2\) design that crosses a minimal image edit with a minimal text edit, yielding 300 headline evaluation cells. Unlike prior unanswerable-VQA benchmarks, \benchmarkname{} tests not only \emph{whether} a question is unanswerable but \emph{why} (via reason codes tied to specific visibility factors), and uses controlled minimal edits to verify that model judgments change when and only when the underlying evidence changes. We score models on confidence-aware accuracy with abstention (CAA), minimal-edit flip rate (MEFR), confidence-ranked selective prediction (SelRank), and second-order perspective reasoning (ToMAcc); all headline numbers are computed on the strict XOR subset (three cells per family, 300 scored items per model). We evaluate nine models spanning flagship and prior-generation closed-source systems, and open-source models from 8B to 12B parameters. GPT-4o and Gemini 3.1 Pro effectively tie for the best composite score (0.728 and 0.727), followed by Gemini 2.5 Pro (0.678). The best open-source model, Gemma 3 12B (0.505), surpasses one prior-generation closed-source system. Text-flip robustness exceeds image-flip robustness for six of nine models, and confidence calibration varies substantially---GPT-4o and Gemini 2.5 Pro achieve similar accuracy yet differ sharply in selective prediction quality.
\end{abstract}

\section{Introduction}

Vision-language models are increasingly deployed in settings where incorrect visual judgments carry real consequences---autonomous driving systems that must detect obscured pedestrians, assistive technologies that describe scenes to blind users, and medical imaging tools that flag ambiguous findings. In all of these domains, a model that guesses when the visual evidence is insufficient can be more dangerous than one that explicitly withholds judgment.

Visibility is a prerequisite for safe and reliable image-grounded reasoning. Many visually phrased questions are unanswerable from a single photo because the relevant evidence is occluded, out of frame, too small, too dark, or not visually observable at all. This paper introduces \benchmarkname, a benchmark designed to quantify whether a system can:
\begin{itemize}[noitemsep, topsep=3pt]
  \item verify simple visibility claims from one image and a short question (testing whether models distinguish ``visible'' from ``present''),
  \item respond appropriately to minimal edits that should flip the correct label (testing robustness to controlled perturbations),
  \item abstain when a human viewer cannot reliably answer from the photo (testing calibrated withholding of judgment).
\end{itemize}

We additionally include a MULTI\_AGENT / SECOND\_ORDER slice that tests second-order perspective judgments---for example, whether the photo supports a claim about what one person can infer about another person's visual access. This capability is essential for collaborative and social reasoning grounded in images.

\paragraph{Setup at a glance.}
Three concepts are needed to read the rest of this paper:
\begin{enumerate}[noitemsep, topsep=3pt]
  \item \textbf{Three labels.} Each prediction is \texttt{VISIBLY\_TRUE}, \texttt{VISIBLY\_FALSE}, or \texttt{ABSTAIN}.
  \item \textbf{\(2\times 2\) family construction.} Every family crosses one minimal image edit with one minimal text edit, producing four cells per family.
  \item \textbf{Three headline cells, one diagnostic.} The composite score uses the first three cells (BASE, TEXT\_FLIP, IMAGE\_FLIP); the fourth cell (DOUBLE\_FLIP) is reported separately as a diagnostic and does not affect the score.
\end{enumerate}

\paragraph{Contributions.}
This paper makes four contributions:
\begin{enumerate}[noitemsep, topsep=3pt]
  \item The \benchmarkname{} benchmark: a task definition, an eight-category visibility taxonomy, and a \(2 \times 2\) family design that crosses minimal image edits with minimal text edits over 100 families.
  \item A metric suite---CAA, MEFR, SelRank, and a weighted composite---tailored to visibility reasoning with explicit abstention and confidence calibration.
  \item An evaluation of nine vision-language models that reveals capability gaps between flagship and open-source systems, an asymmetry between text-flip and image-flip robustness, and substantial variation in confidence calibration.
  \item Public release of the full dataset, per-image metadata, and evaluation infrastructure.
\end{enumerate}

\paragraph{Scope.}
\benchmarkname{} is not a general VQA benchmark. It deliberately isolates a single skill---deciding whether a short visibility claim is supported by the pixels in one photograph---and measures that skill under controlled perturbations. The tight scope is by design: it enables the \(2\times 2\) construction that makes flip-rate and calibration analyses meaningful, at the cost of not covering open-ended reasoning, multi-image comparison, or long-form answers.

\paragraph{Data release.}
The full dataset, metadata, and evaluation code are released at the repository linked below.\footnote{\url{https://github.com/neilt93/Paper-with-Davis}}

\section{Qualitative examples}
\label{sec:examples}

\begin{figure}[p]
  \centering
  \scriptsize
  \captionsetup{font=scriptsize}
  \caption{Representative base/flip examples (one family per primary category). Each pair shows the BASE cell \((I^0,q^0)\) and IMAGE\_FLIP cell \((I^1,q^0)\). Most pairs follow the strict XOR pattern: BASE is \texttt{VF} and IMAGE\_FLIP is \texttt{VT}. The IC pair illustrates \texttt{ABSTAIN}. (\texttt{VT}=\texttt{VISIBLY\_TRUE}, \texttt{VF}=\texttt{VISIBLY\_FALSE}.)}
  \label{fig:category_examples}

  % Row 1: LD and OF
\begin{subfigure}[t]{0.24\textwidth}
  \centering
  \maybeincludegraphics[height=0.15\textheight,keepaspectratio]{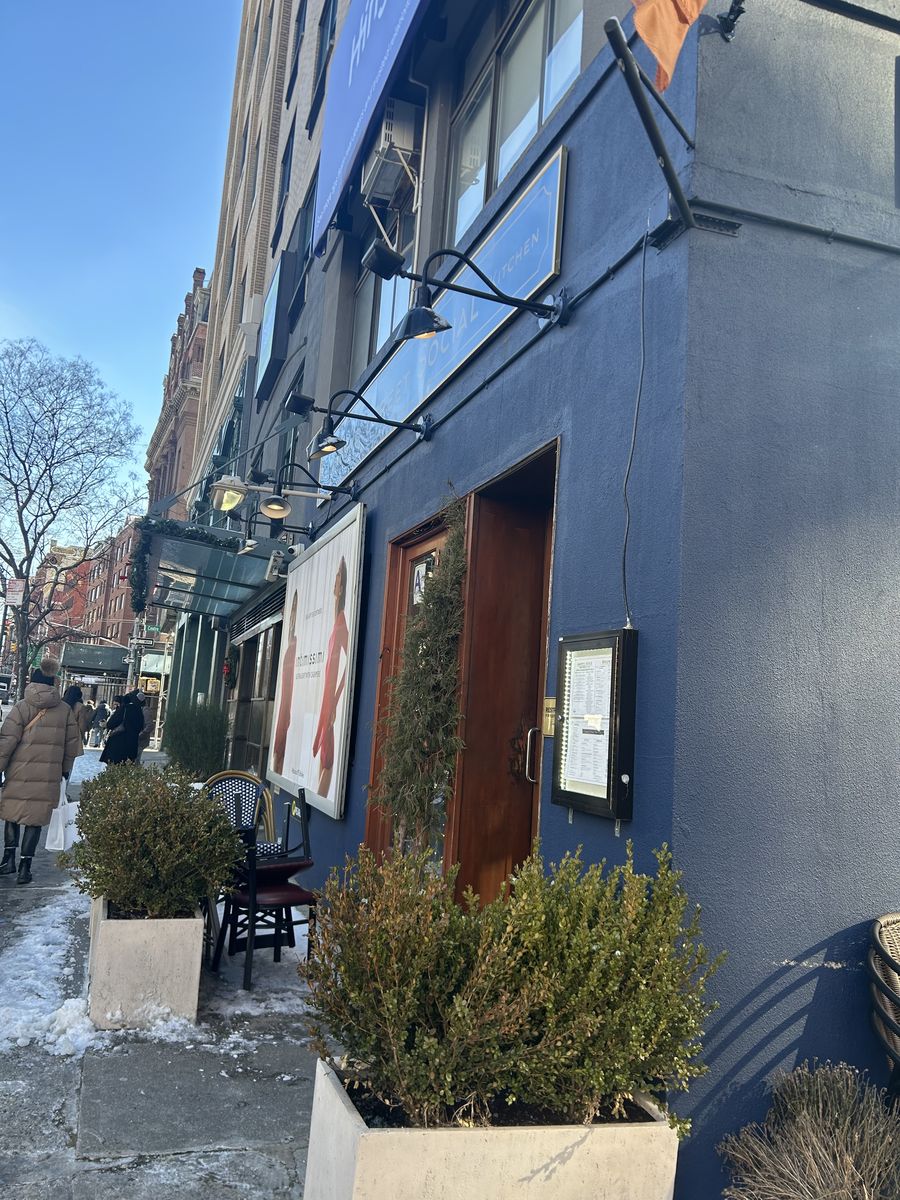}{\linewidth}
  \caption{\textbf{LD-02 BASE.}\\
  \textbf{Q:} ``Is the sign text clearly readable in this photo?''\\
  \textbf{Gold:} \texttt{VISIBLY\_FALSE}.}
\end{subfigure}\hfill
\begin{subfigure}[t]{0.24\textwidth}
  \centering
  \maybeincludegraphics[height=0.15\textheight,keepaspectratio]{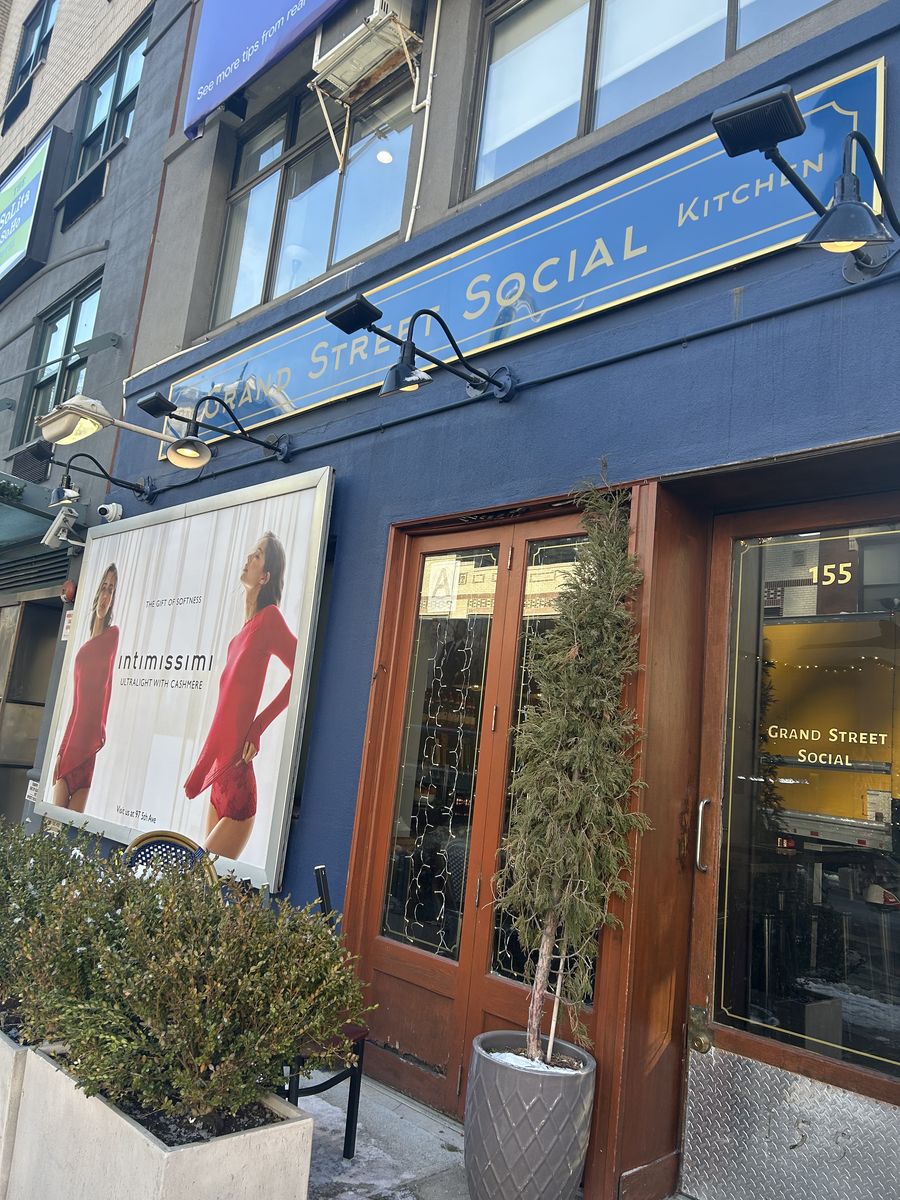}{\linewidth}
  \caption{\textbf{LD-02 IMAGE\_FLIP.}\\
  \textbf{Gold:} \texttt{VISIBLY\_TRUE}.}
\end{subfigure}\hfill
\begin{subfigure}[t]{0.24\textwidth}
  \centering
  \maybeincludegraphics[height=0.15\textheight,keepaspectratio]{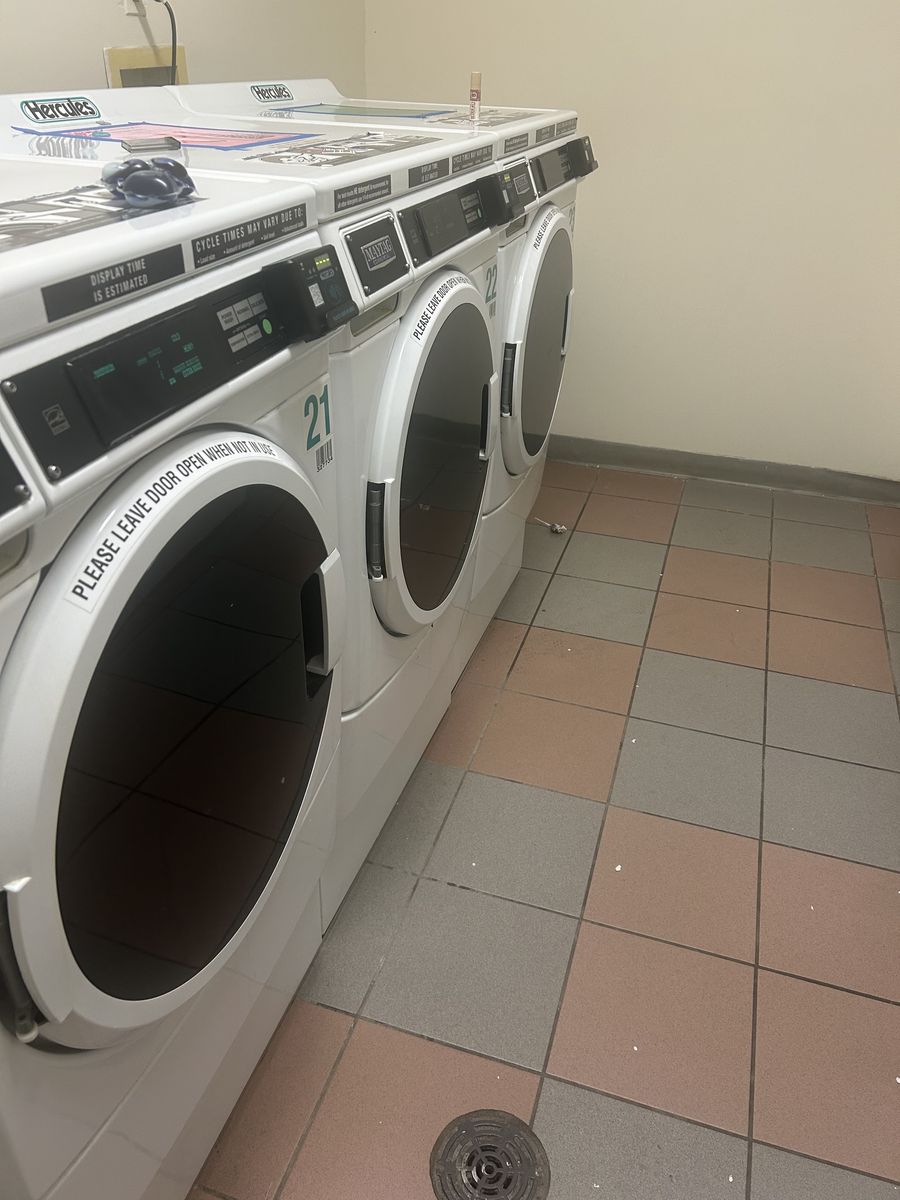}{\linewidth}
  \caption{\textbf{OF-13 BASE.}\\
  \textbf{Q:} ``Is the laundry basket visible in this photo?''\\
  \textbf{Gold:} \texttt{VISIBLY\_FALSE}.}
\end{subfigure}\hfill
\begin{subfigure}[t]{0.24\textwidth}
  \centering
  \maybeincludegraphics[height=0.15\textheight,keepaspectratio]{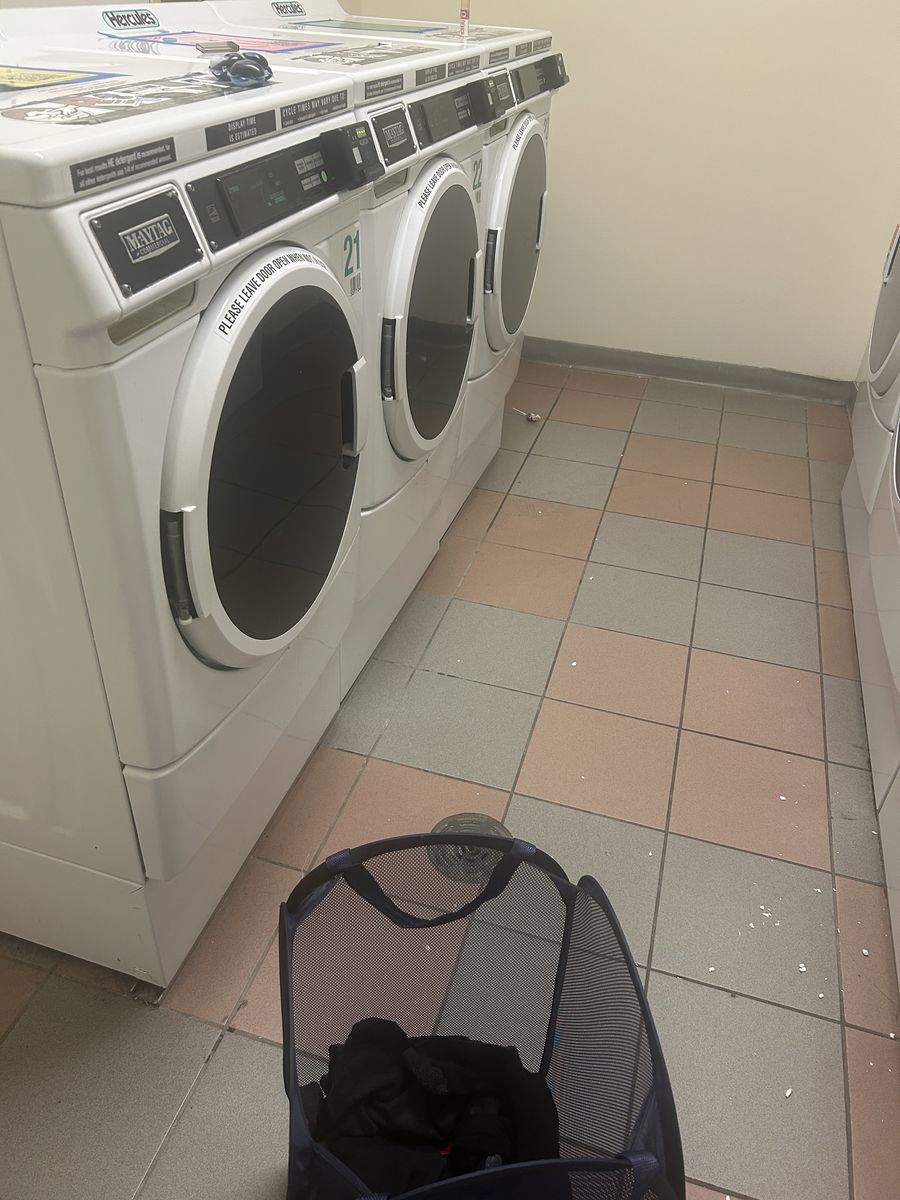}{\linewidth}
  \caption{\textbf{OF-13 IMAGE\_FLIP.}\\
  \textbf{Gold:} \texttt{VISIBLY\_TRUE}.}
\end{subfigure}

\vspace{2pt}

% Row 2: OC and GD
\begin{subfigure}[t]{0.24\textwidth}
  \centering
  \maybeincludegraphics[height=0.15\textheight,keepaspectratio]{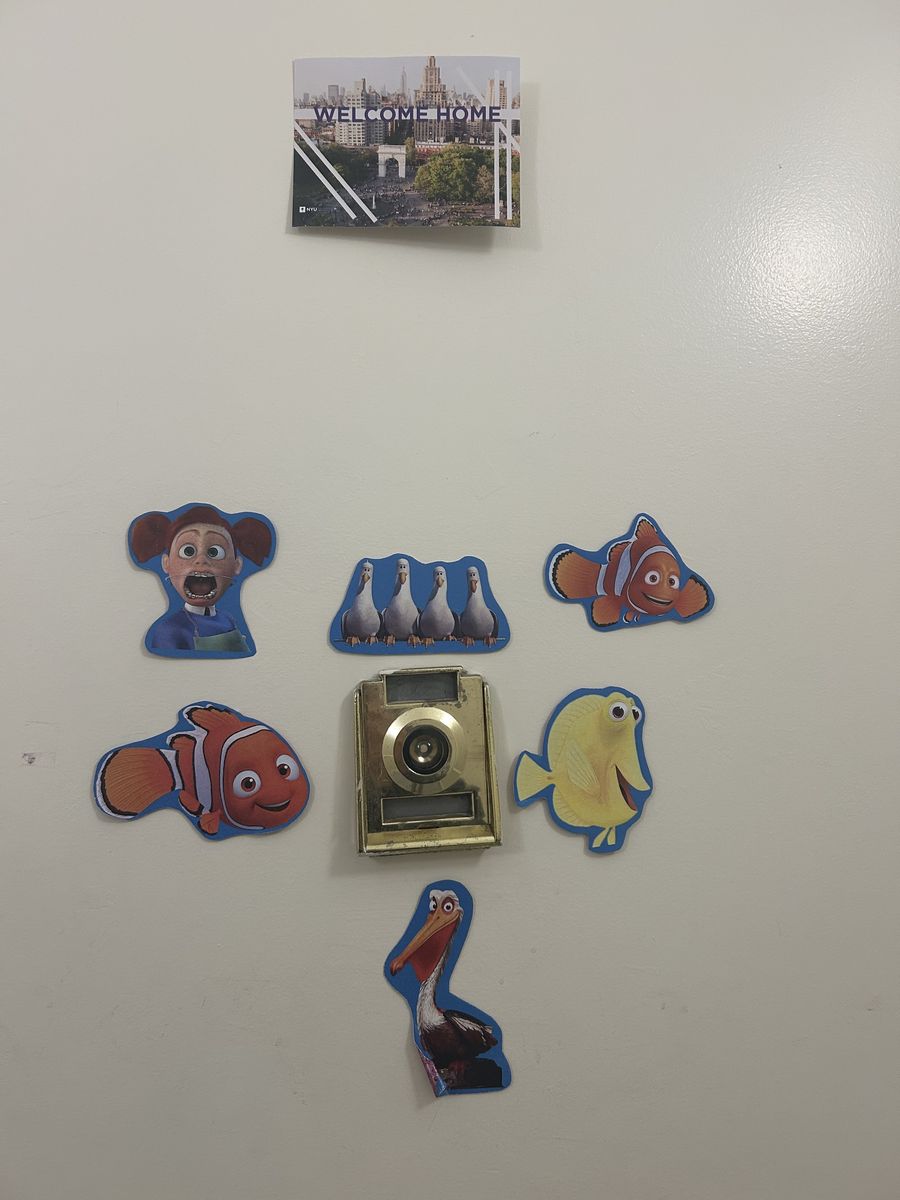}{\linewidth}
  \caption{\textbf{OC-11 BASE.}\\
  \textbf{Q:} ``Is the room number clearly visible in this photo?''\\
  \textbf{Gold:} \texttt{VISIBLY\_FALSE}.}
\end{subfigure}\hfill
\begin{subfigure}[t]{0.24\textwidth}
  \centering
  \maybeincludegraphics[height=0.15\textheight,keepaspectratio]{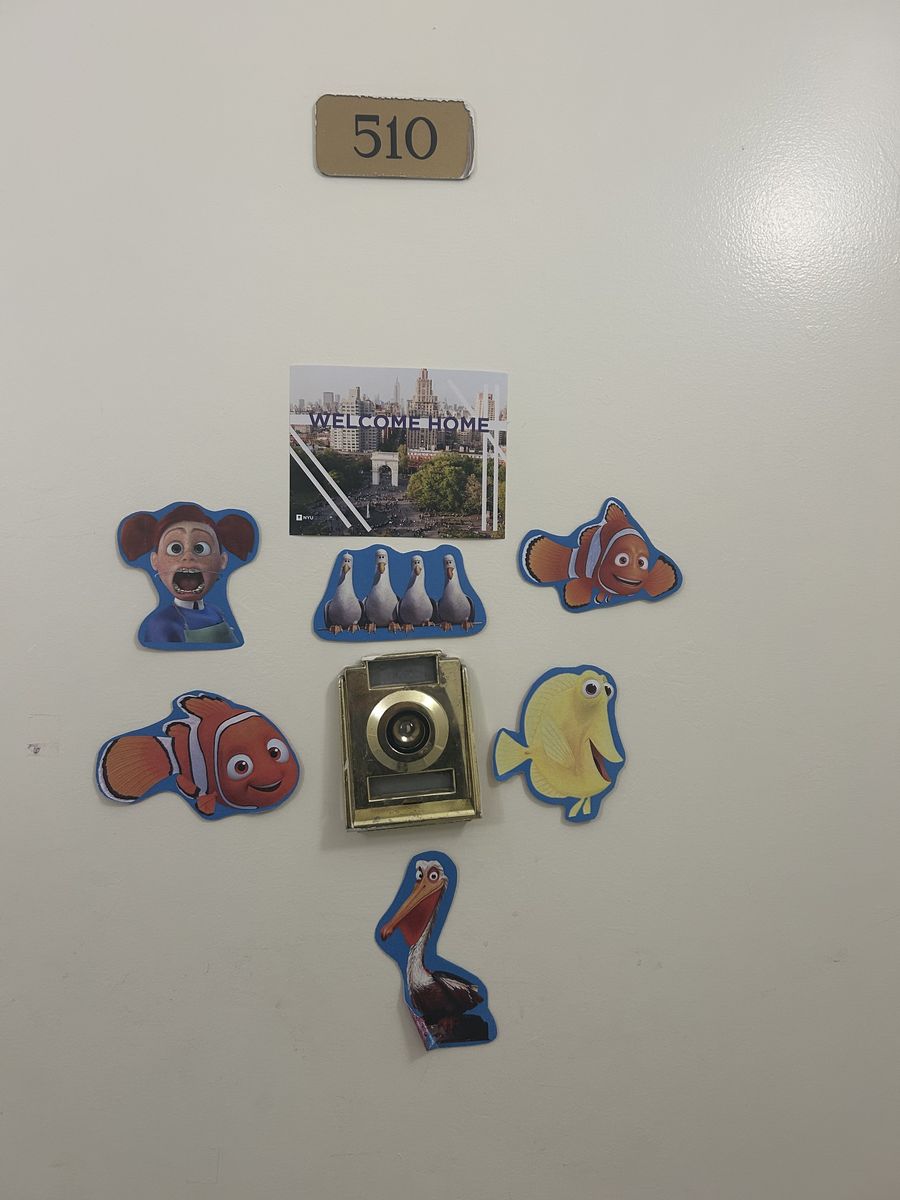}{\linewidth}
  \caption{\textbf{OC-11 IMAGE\_FLIP.}\\
  \textbf{Gold:} \texttt{VISIBLY\_TRUE}.}
\end{subfigure}\hfill
\begin{subfigure}[t]{0.24\textwidth}
  \centering
  \maybeincludegraphics[height=0.15\textheight,keepaspectratio]{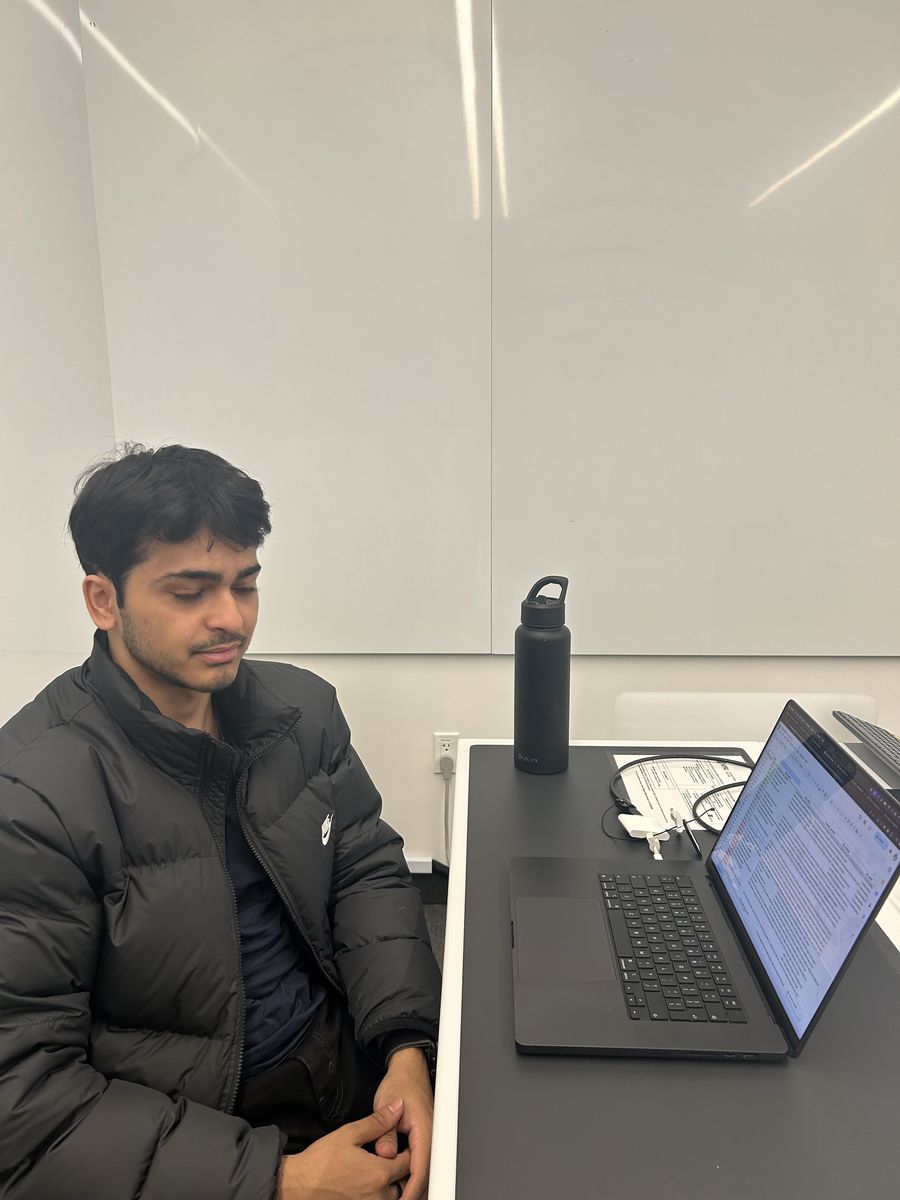}{\linewidth}
  \caption{\textbf{GD-06 BASE.}\\
  \textbf{Q:} ``Is he clearly looking at the laptop screen?''\\
  \textbf{Gold:} \texttt{VISIBLY\_FALSE}.}
\end{subfigure}\hfill
\begin{subfigure}[t]{0.24\textwidth}
  \centering
  \maybeincludegraphics[height=0.15\textheight,keepaspectratio]{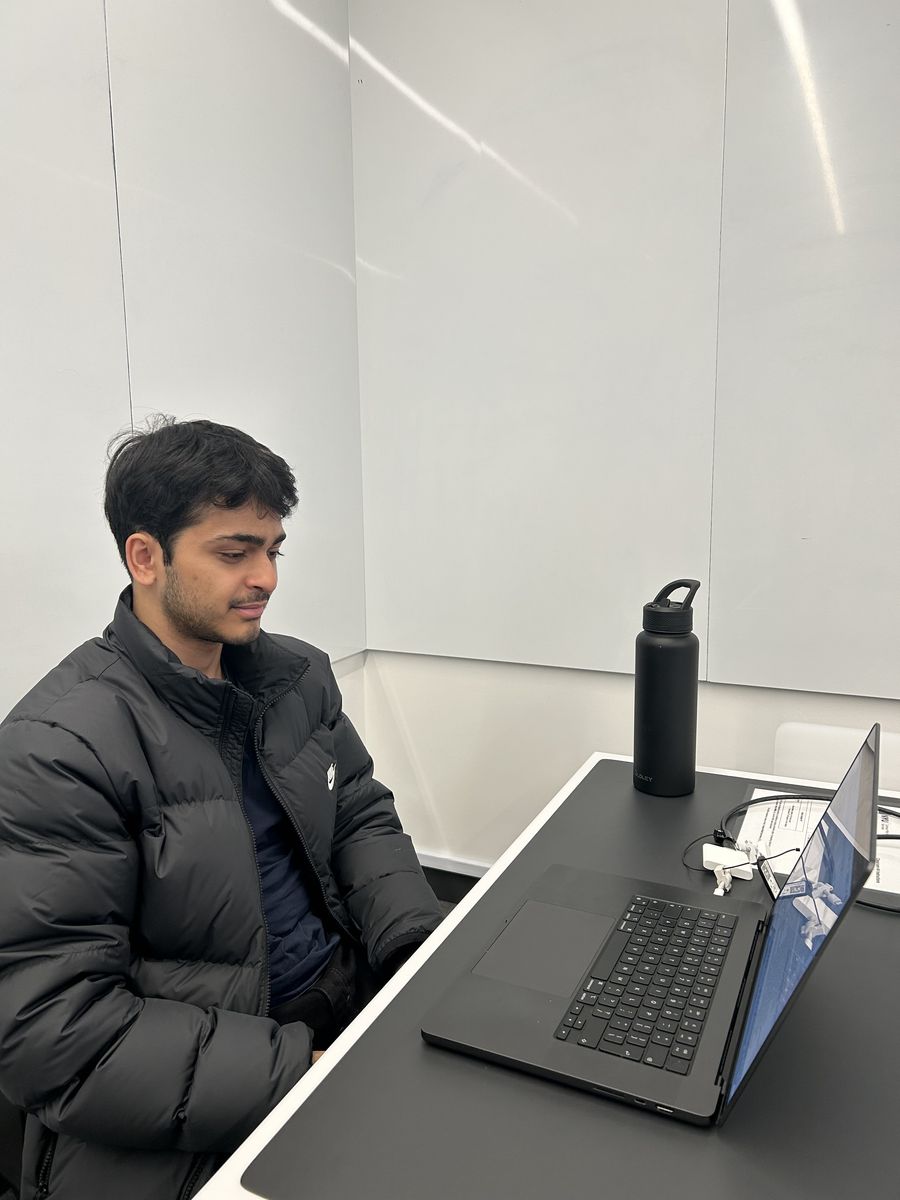}{\linewidth}
  \caption{\textbf{GD-06 IMAGE\_FLIP.}\\
  \textbf{Gold:} \texttt{VISIBLY\_TRUE}.}
\end{subfigure}

\vspace{2pt}

% Row 3: AV and NV
\begin{subfigure}[t]{0.24\textwidth}
  \centering
  \maybeincludegraphics[height=0.15\textheight,keepaspectratio]{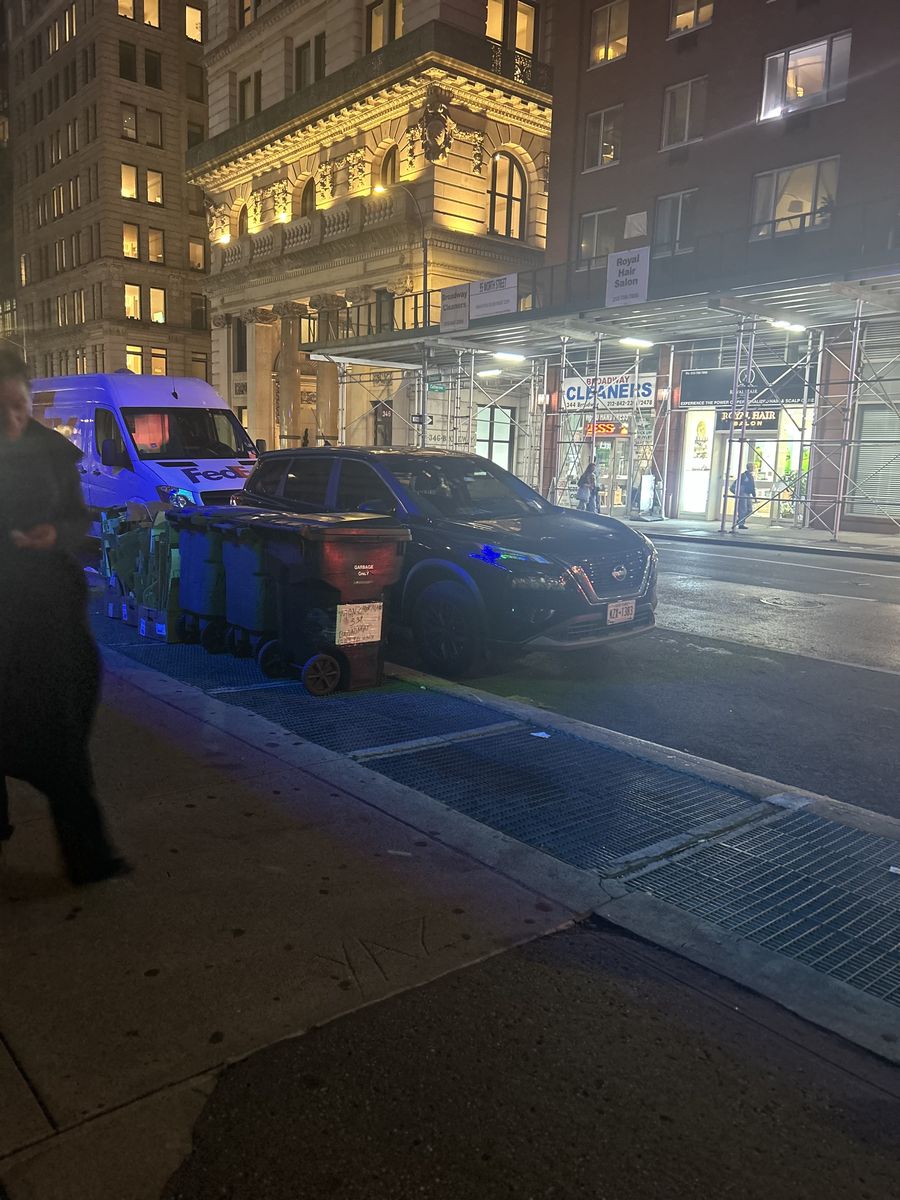}{\linewidth}
  \caption{\textbf{AV-01 BASE.}\\
  \textbf{Q:} ``Is the building number clearly readable in this photo?''\\
  \textbf{Gold:} \texttt{VISIBLY\_FALSE}.}
\end{subfigure}\hfill
\begin{subfigure}[t]{0.24\textwidth}
  \centering
  \maybeincludegraphics[height=0.15\textheight,keepaspectratio]{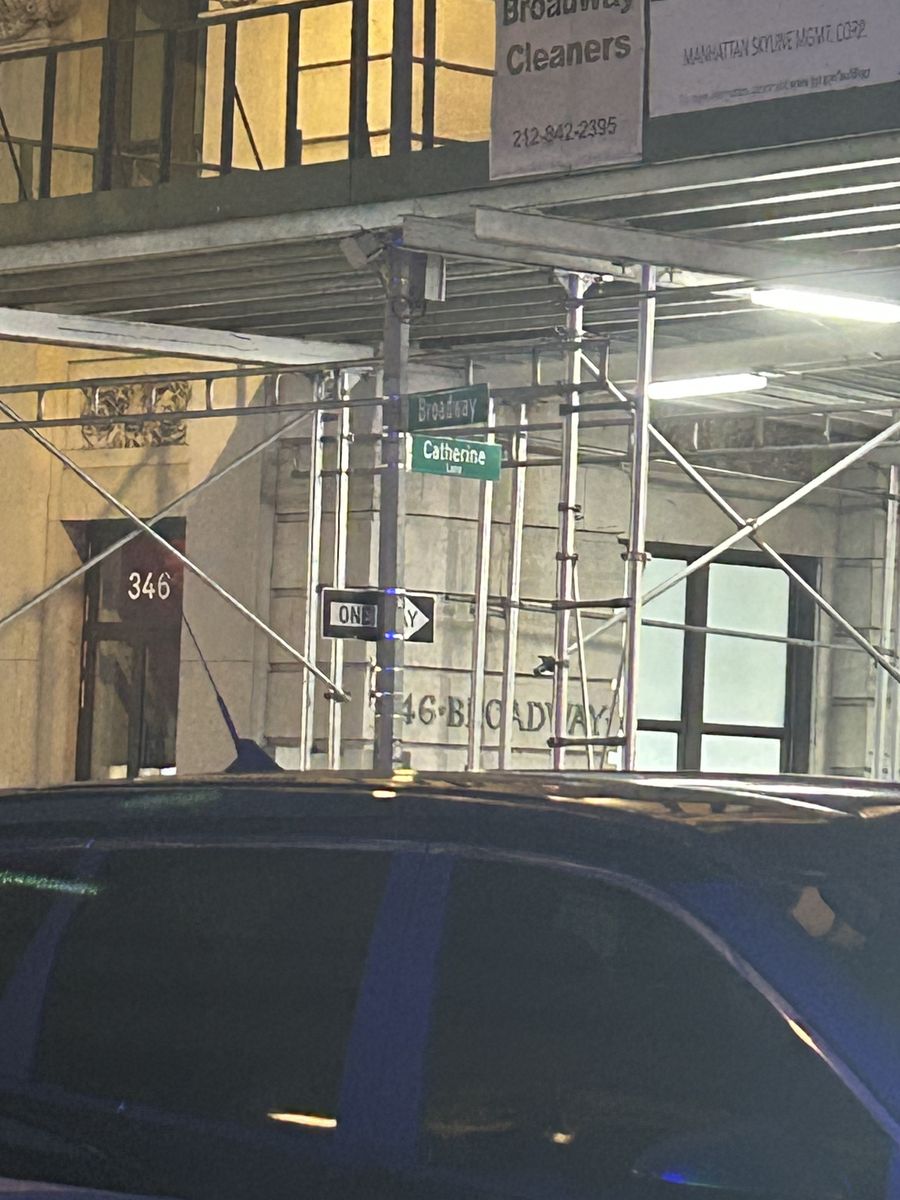}{\linewidth}
  \caption{\textbf{AV-01 IMAGE\_FLIP.}\\
  \textbf{Gold:} \texttt{VISIBLY\_TRUE}.}
\end{subfigure}\hfill
\begin{subfigure}[t]{0.24\textwidth}
  \centering
  \maybeincludegraphics[height=0.15\textheight,keepaspectratio]{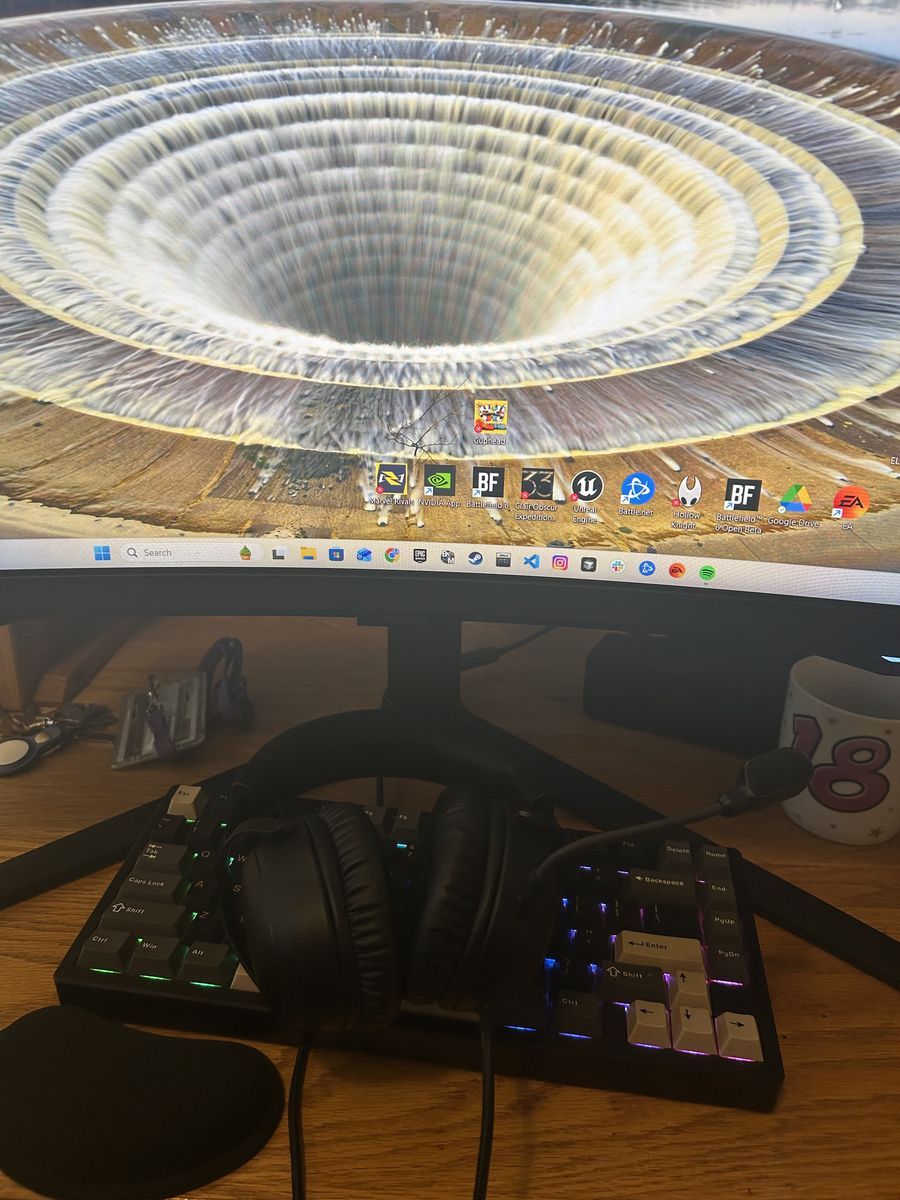}{\linewidth}
  \caption{\textbf{NV-08 BASE.}\\
  \textbf{Q:} ``Is the song title/track info visible in this photo (screen/now playing)?''\\
  \textbf{Gold:} \texttt{VISIBLY\_FALSE}.}
\end{subfigure}\hfill
\begin{subfigure}[t]{0.24\textwidth}
  \centering
  \maybeincludegraphics[height=0.15\textheight,keepaspectratio]{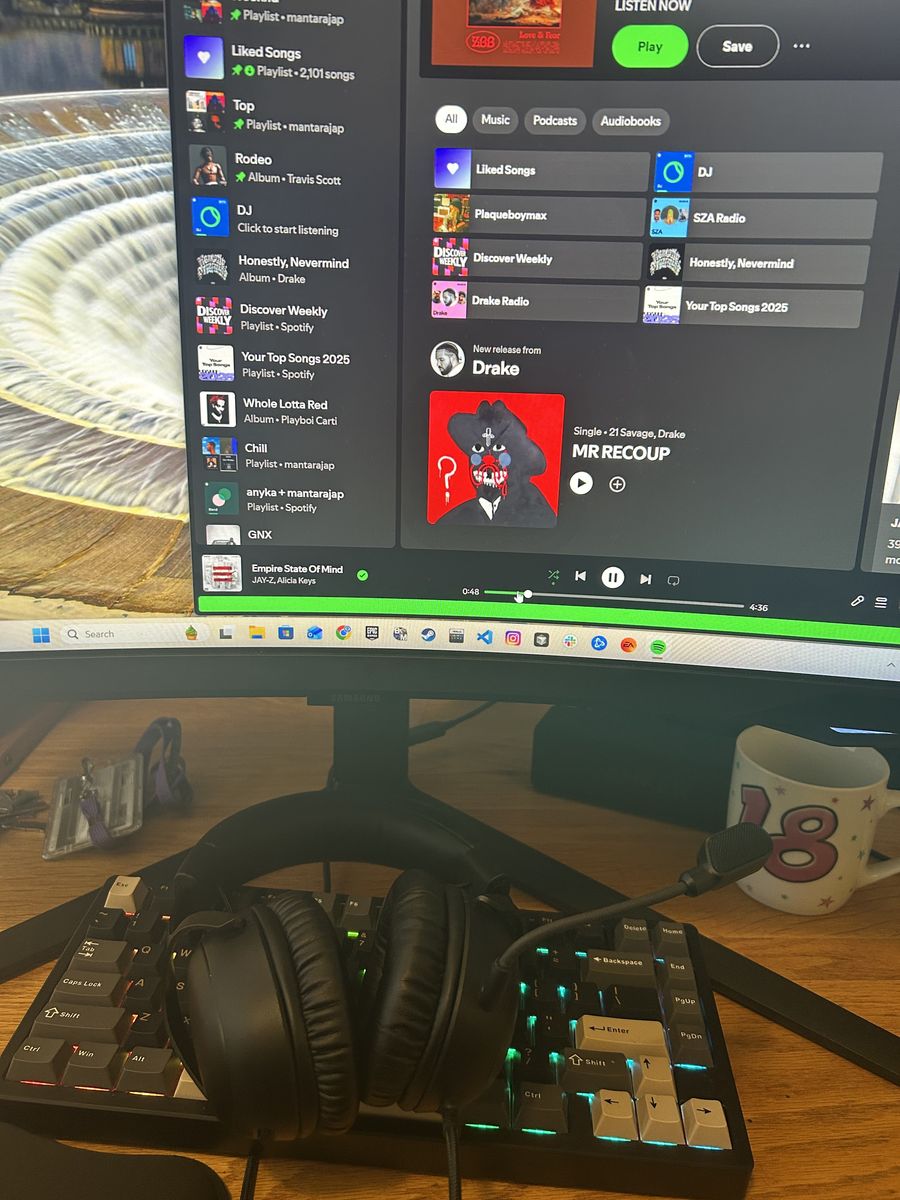}{\linewidth}
  \caption{\textbf{NV-08 IMAGE\_FLIP.}\\
  \textbf{Gold:} \texttt{VISIBLY\_TRUE}.}
\end{subfigure}

\vspace{2pt}

% Row 4: IC and MA/SO
\begin{subfigure}[t]{0.24\textwidth}
  \centering
  \maybeincludegraphics[height=0.15\textheight,keepaspectratio]{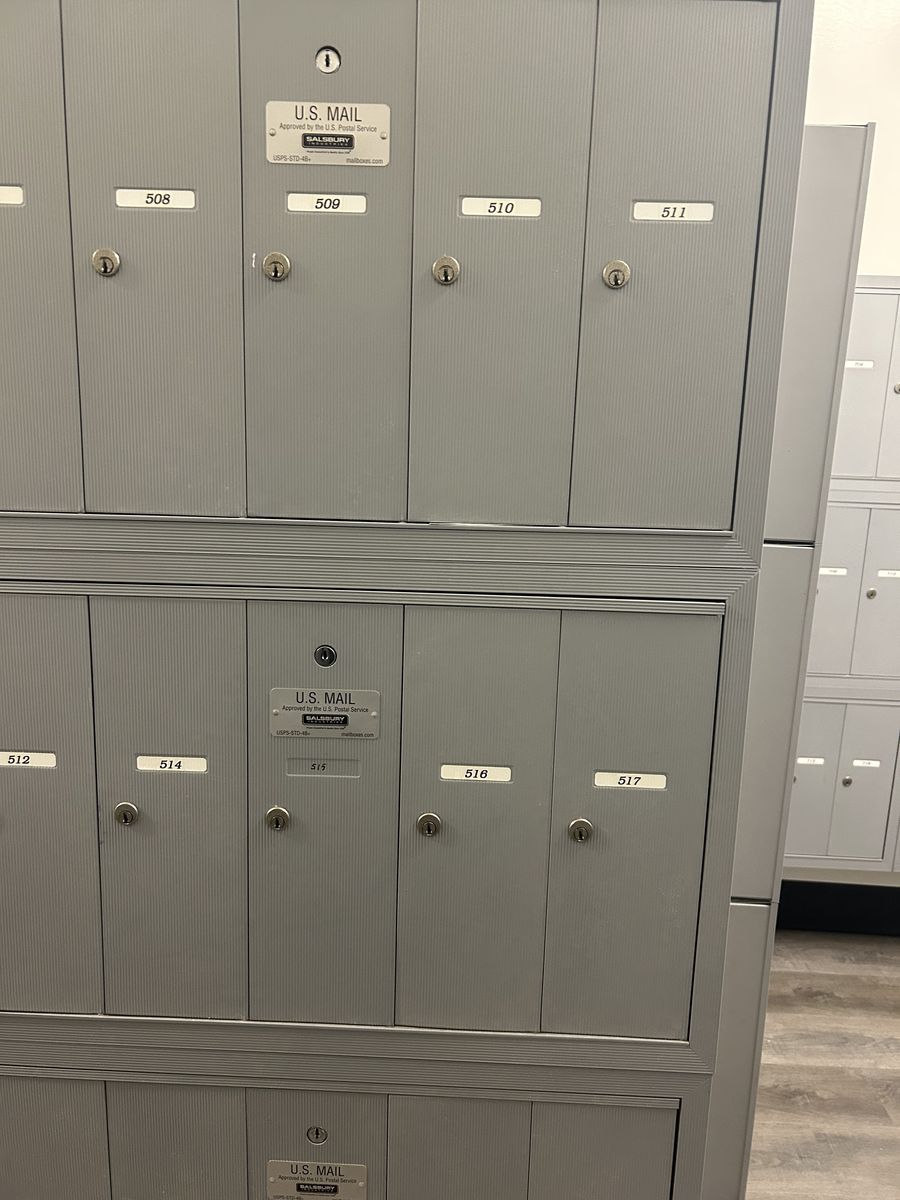}{\linewidth}
  \caption{\textbf{IC-05 BASE.}\\
  \textbf{Q:} ``Is the correct mailbox clearly identifiable in this photo?''\\
  \textbf{Gold:} \texttt{ABSTAIN}.}
\end{subfigure}\hfill
\begin{subfigure}[t]{0.24\textwidth}
  \centering
  \maybeincludegraphics[height=0.15\textheight,keepaspectratio]{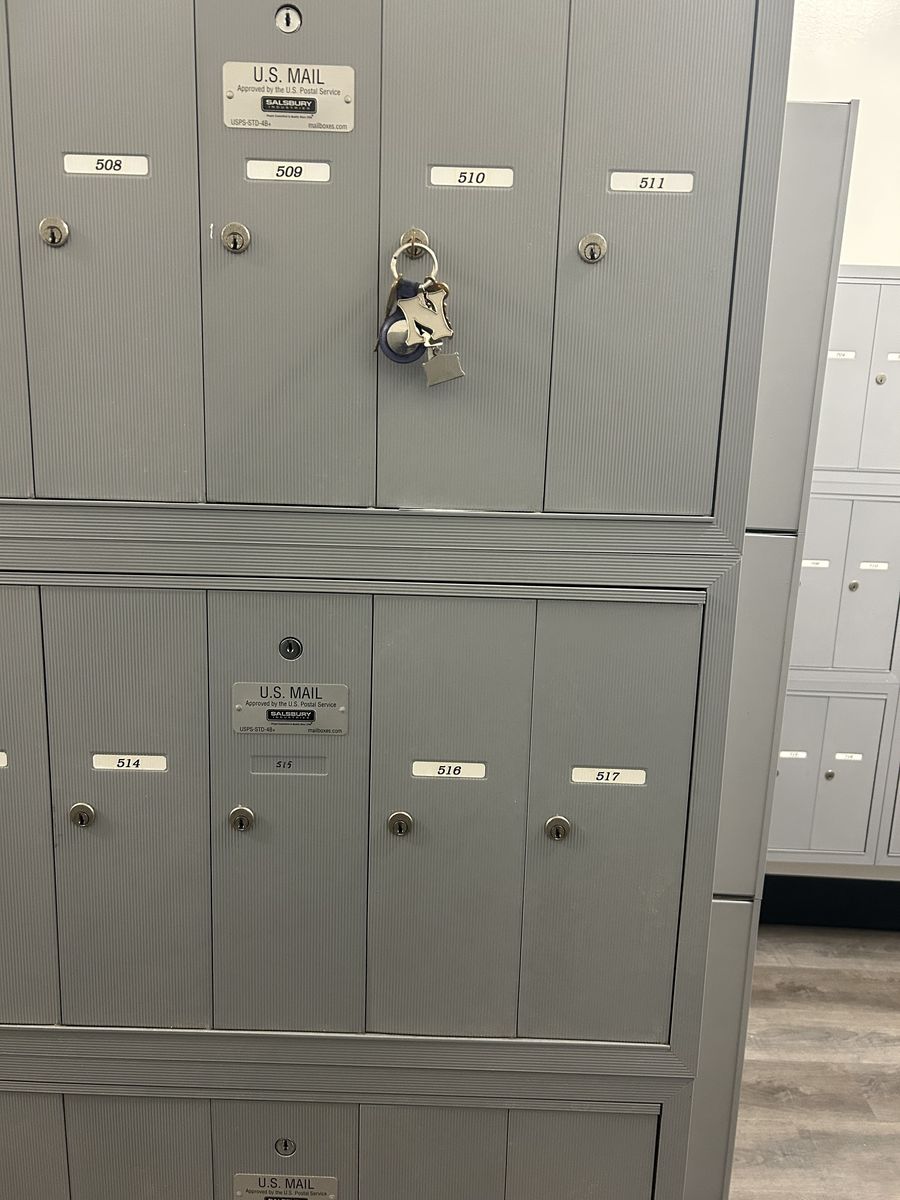}{\linewidth}
  \caption{\textbf{IC-05 IMAGE\_FLIP.}\\
  \textbf{Gold:} \texttt{VISIBLY\_TRUE}.}
\end{subfigure}\hfill
\begin{subfigure}[t]{0.24\textwidth}
  \centering
  \maybeincludegraphics[height=0.15\textheight,keepaspectratio]{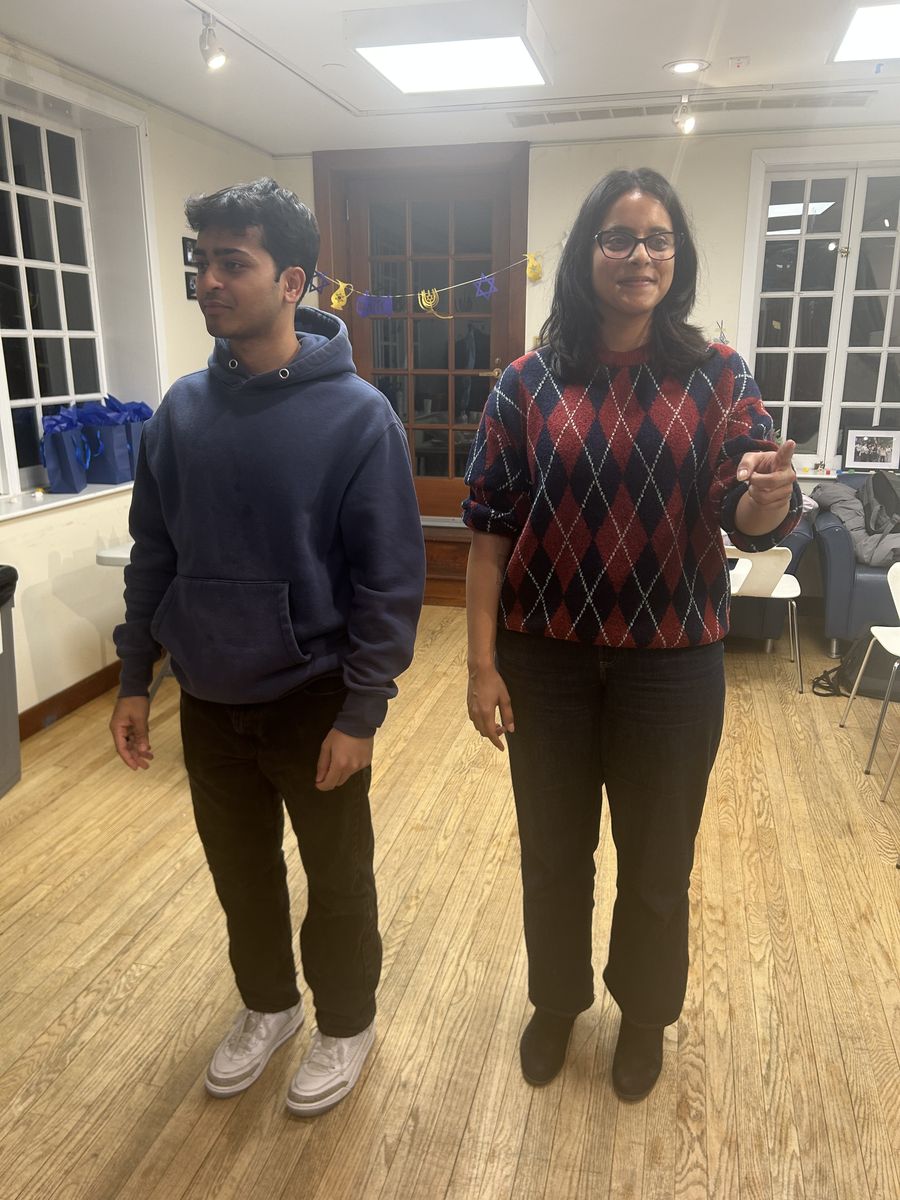}{\linewidth}
  \caption{\textbf{MA-06 BASE.}\\
  \textbf{Q:} ``Is he clearly looking at what she is pointing to?''\\
  \textbf{Gold:} \texttt{VISIBLY\_FALSE}.}
\end{subfigure}\hfill
\begin{subfigure}[t]{0.24\textwidth}
  \centering
  \maybeincludegraphics[height=0.15\textheight,keepaspectratio]{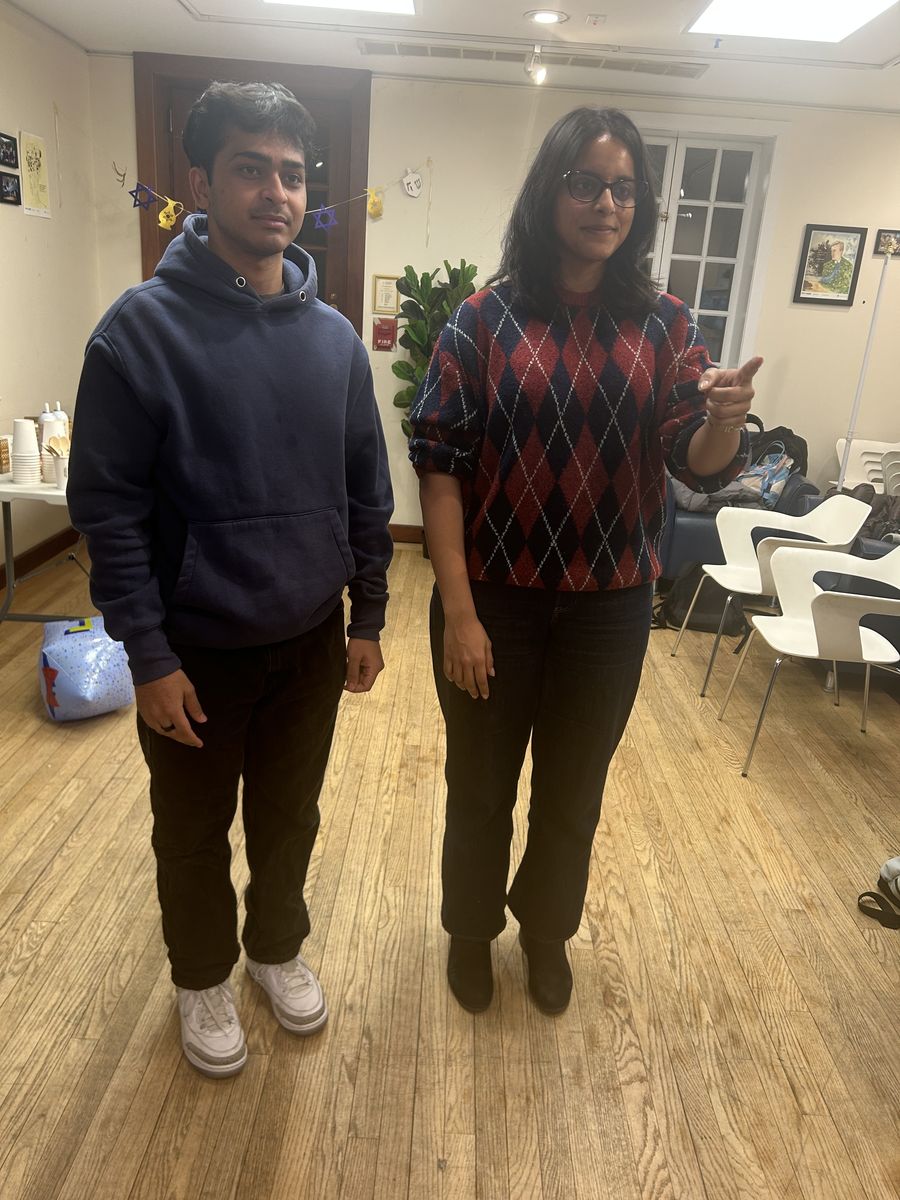}{\linewidth}
  \caption{\textbf{MA-06 IMAGE\_FLIP.}\\
  \textbf{Gold:} \texttt{VISIBLY\_TRUE}.}
\end{subfigure}

\end{figure}

Figure~\ref{fig:category_examples} shows one representative base/flip pair per primary category. Each pair uses the base question \(q^0\) and displays the base image \(I^0\) and the minimally edited image \(I^1\) (the IMAGE\_FLIP cell). For space, we omit the corresponding text-edited question \(q^1\) and the DOUBLE\_FLIP cell, but every family in the released benchmark includes all four cells \((I^a,q^b)\) for \(a,b \in \{0,1\}\) (Section~\ref{subsec:xor}).

\paragraph{Note on \texttt{ABSTAIN} examples.}
\texttt{ABSTAIN} is a valid gold label in \benchmarkname{} when a careful human cannot decide. The main headline results in Section~\ref{sec:results} focus on the strict XOR headline subset used for score aggregation. \texttt{ABSTAIN}-labelled items are included for completeness and analysis, and we recommend reporting their separate accuracy and abstention rates in future releases.

\section{Related work}
\label{sec:related}

\paragraph{Unanswerable visual questions and withholding judgment.}
Davis points out that many questions about images are inherently unanswerable even with perfect vision because the relevant facts can be occluded, outside the frame, or non-visual \citep{davis_unanswerable_images}. Several benchmarks highlight that real images often do not support a definitive answer. VizWiz collects question--image pairs from blind or vision-impaired users, many of which lack sufficient visual evidence for a confident answer \citep{gurari_vizwiz_2018}. Recent work explicitly evaluates abstention on unanswerable visual questions, for example UNK-VQA \citep{guo_unkvqa_2023} and TUBench \citep{he_tubench_2024}. \benchmarkname{} differs from these efforts in two ways: it tests not just \emph{whether} a question is unanswerable but \emph{why} (via reason codes tied to specific visibility factors), and it uses controlled minimal edits to verify that models change their judgments when and only when the underlying evidence changes.

\paragraph{Visibility factors: gaze, occlusion, and field of view.}
Datasets such as GazeFollow \citep{recasens_gazefollow_2015} and Following Gaze Across Views \citep{recasens_following_gaze_2017} study gaze targets and off-frame gaze. Work on occlusion reasoning analyzes how blockers and partial views affect recognition and localization \citep{jiang_peekaboo_2020}. \benchmarkname{} differs by evaluating label-level support for a visibility claim under controlled edits rather than requiring precise localization.

\paragraph{Hallucination and faithfulness in vision-language models.}
Benchmarks such as MME \citep{fu_mme_2023} and POPE \citep{li_pope_2023} evaluate multimodal perception, cognition, and hallucinated object claims. \benchmarkname{} complements these by isolating a narrower but safety-relevant skill: deciding whether a claim is supported by visible evidence, and withholding judgment when it is not.

\paragraph{Selective prediction, reject options, and risk-coverage.}
Selective classification formalizes the trade-off between risk and coverage when a classifier may reject \citep{elyaniv_selective_2010}. SelectiveNet and follow-up work evaluate selective prediction using risk-coverage curves and related summary statistics \citep{geifman_selectivenet_2019,varshney_selective_2022}. \benchmarkname{} uses a confidence-ranked selective prediction score to test whether model confidence aligns with correctness among answered items.

\paragraph{Second-order perspective and theory-of-mind style probes.}
Recent work motivates stress tests for second-order perspective judgments and their brittleness under structured perturbations \citep{kosinski_tom_llms_2023,ullman_tom_perturbations}. \benchmarkname{} includes a dedicated MULTI\_AGENT / SECOND\_ORDER slice in which the label depends on what the photo supports about one agent's knowledge of another agent's visual access. Unlike text-only theory-of-mind probes, \benchmarkname{} requires that second-order judgments be grounded in image evidence---the model must reason about what is visually accessible to each agent depicted in the photograph.

\section{Benchmark design}
\label{sec:design}

\subsection{Task definition and labels}

Each item consists of an image and a short yes/no question that expresses a \emph{visibility claim}. The system's job is to decide whether that claim is supported by visible evidence in the photo.

The label space is:
\begin{itemize}[noitemsep, topsep=3pt]
  \item \texttt{VISIBLY\_TRUE}: the claim is supported by visible evidence, defined as ``a careful human would answer \emph{yes} from this photo with reasonable confidence'';
  \item \texttt{VISIBLY\_FALSE}: the claim is contradicted by the photo, defined as ``a careful human would answer \emph{no} from this photo with reasonable confidence'';
  \item \texttt{ABSTAIN}: the photo does not support either a confident yes or a confident no.
\end{itemize}

The labels always refer to whether the \emph{question's claim} is supported. This allows both positive and negative question forms, for example ``Is the serial number readable?'' versus ``Is the serial number unreadable?''

\subsection{Structured outputs}

Systems output a single structured prediction:
\begin{itemize}[noitemsep, topsep=3pt]
  \item \texttt{label} \(\in \{\texttt{VISIBLY\_TRUE}, \texttt{VISIBLY\_FALSE}, \texttt{ABSTAIN}\}\),
  \item \texttt{reason\_code} \(\in \{\)GAZE\_DIRECTION, OCCLUSION, OUT\_OF\_FRAME, LIGHTING\_DISTANCE, INHERENTLY\_NONVISUAL, AUGMENTED\_VISION\_REQUIRED, INSUFFICIENT\_CONTEXT, MULTI\_AGENT\_SECOND\_ORDER, NONE\(\}\),
  \item \texttt{confidence} \(\in [0,1]\), interpreted as the model's probability that its chosen label is correct.
\end{itemize}

\paragraph{Reason-code conventions.}
If \texttt{label}=\texttt{VISIBLY\_TRUE}, set \texttt{reason\_code}=\texttt{NONE}. If \texttt{label}=\texttt{ABSTAIN}, choose exactly one limiting-factor code. If \texttt{label}=\texttt{VISIBLY\_FALSE}, choose a limiting-factor code when the claim is false due to a visibility limitation (for example, ``Is the serial number readable?'' when the serial number is present but blurred). If the claim is directly refuted with no visibility limitation required (for example, ``Is the mug not visible?'' when the mug is plainly visible), set \texttt{reason\_code}=\texttt{NONE}.

\paragraph{Why include \texttt{reason\_code}?}
Reason codes make failures more actionable and more interpretable. They help distinguish ``the evidence is missing'' (out of frame) from ``the evidence is present but blocked'' (occlusion) or ``present but too small'' (distance), which correspond to different corrective actions (change viewpoint, remove blocker, move closer, increase illumination, or gather context).

\paragraph{Choosing a single code.}
If multiple limiting factors apply, we use the following precedence to select one:
\begin{quote}
\begin{lstlisting}[basicstyle=\small\ttfamily]
OCCLUSION > OUT_OF_FRAME > GAZE_DIRECTION
> LIGHTING_DISTANCE > AUGMENTED_VISION_REQUIRED
> INHERENTLY_NONVISUAL > INSUFFICIENT_CONTEXT
> MULTI_AGENT_SECOND_ORDER
\end{lstlisting}
\end{quote}
The same precedence is included in the evaluation prompt in Section~\ref{sec:prompt}.

\subsection{Family structure: \(2 \times 2\) design}

Items are grouped into families. Each family is built around:
\begin{itemize}[noitemsep, topsep=3pt]
  \item a base image \(I^0\) and an edited image \(I^1\) that differs by one atomic scene change,
  \item a base question \(q^0\) and a text-edited question \(q^1\) that flips the underlying claim being tested.
\end{itemize}

This yields four evaluated cells \((I^a, q^b)\) for \(a,b \in \{0,1\}\):
\[
(I^0,q^0),\ (I^0,q^1),\ (I^1,q^0),\ (I^1,q^1).
\]

\subsection{Atomic scene changes (minimal image edits)}
\label{subsec:atomic}

An \emph{atomic scene change} is a minimal change to the photographed world intended to affect exactly one visibility factor relevant to the family claim, while leaving the remainder of the scene as stable as possible. In practice, families were re-shot rather than digitally edited, and the atomic change was enacted physically (for example, moving closer to a screen, shifting an occluder, or moving a target slightly into frame).

We aim for the following constraints:
\begin{itemize}[noitemsep, topsep=3pt]
  \item \textbf{Single causal factor:} the edit targets one visibility factor (for example occlusion or distance) for the referent used in the question.
  \item \textbf{No incidental cues:} no new salient objects are introduced unless they are the edit itself (for example the occluder being moved).
  \item \textbf{Stable viewpoint:} camera position and framing are held fixed unless the family tests OUT\_OF\_FRAME or LIGHTING\_DISTANCE and requires a controlled viewpoint change.
  \item \textbf{Stable lighting:} lighting conditions are held fixed unless the family targets LIGHTING\_DISTANCE.
\end{itemize}

The DOUBLE\_FLIP cell (Section~\ref{subsec:xor}) provides a diagnostic for unintended interactions, since composing the text and image edits should re-invert the claim under the intended construction.

\subsection{XOR construction and the diagnostic fourth cell}
\label{subsec:xor}

For the strict headline subset used in the main score, families are constructed so that the gold labels follow a fixed XOR pattern over the two edits:
\[
y^{00}=\texttt{VISIBLY\_FALSE},\quad
y^{01}=\texttt{VISIBLY\_TRUE},\quad
y^{10}=\texttt{VISIBLY\_TRUE},\quad
y^{11}=\texttt{VISIBLY\_FALSE}.
\]
Intuitively, the base cell is designed to be confidently refuted by the photo, while either a text edit alone or an image edit alone makes the claim supported. When both edits are applied, the claim is again refuted.

We treat the first three cells as \textbf{headline cells} for the main score:
\[
(I^0,q^0)\ \text{BASE},\quad (I^0,q^1)\ \text{TEXT\_FLIP},\quad (I^1,q^0)\ \text{IMAGE\_FLIP}.
\]
The fourth cell \((I^1,q^1)\) (\textbf{DOUBLE\_FLIP}) is \textbf{diagnostic only}. We report it separately and do not include it in the composite headline score.

\subsection{Primary categories}

\benchmarkname{} is organized around mutually exclusive primary visibility factors at the family level. Each base image is tagged with exactly one primary category.

\begin{table}[t]
  \centering
  \small
  \setlength{\tabcolsep}{5pt}
  \caption{Primary category label on each scene, what it tests, an example prompt (written to avoid encoding the category in the question text), and the number of base pictures in the current release.}
  \label{tab:categories}
  \resizebox{\linewidth}{!}{%
  \begin{tabular}{l p{3.2cm} p{4.6cm} S[table-format=3.0]}
    \toprule
    \textbf{Category (primary label on scene)} & \textbf{What it tests} & \textbf{Example prompt (short)} & \textbf{\# Base Pictures} \\
    \midrule
    GAZE\_DIRECTION & Head and eye orientation versus target & ``Is Pat looking at the mug?'' & 20 \\
    OCCLUSION & Opaque blockers along line of sight & ``Is the metal key blade visible?'' & 20 \\
    OUT\_OF\_FRAME & Target outside current view or crop & ``Is the dog visible in the photo?'' & 15 \\
    LIGHTING\_DISTANCE & Darkness, glare, or distance limits & ``Is the sign text readable?'' & 13 \\
    INHERENTLY\_NONVISUAL & Properties vision cannot reveal & ``Is the device PIN visible anywhere?'' & 11 \\
    AUGMENTED\_VISION\_REQUIRED & Requires magnification not available from the base photo & ``Is the fine print readable?'' & 7 \\
    INSUFFICIENT\_CONTEXT & Under-specified referents or missing scene facts & ``Is the correct key visible?'' & 7 \\
    MULTI\_AGENT / SECOND\_ORDER & Second-order perspective judgments grounded in one photo & ``Does Bob know Alice cannot see the card?'' & 7 \\
    \midrule
    \textbf{TOTAL} & & & 100 \\
    \bottomrule
  \end{tabular}%
  }
\end{table}

\subsection{Multi-agent and second-order visibility}
\label{subsec:multiagent}

The MULTI\_AGENT / SECOND\_ORDER subset targets second-order judgments, where the system must decide whether the photo supports a claim about one agent's knowledge of another agent's visual access.

\paragraph{Writing rule (referents must be clear).}
SECOND\_ORDER questions name agents and targets explicitly (for example ``Bob'', ``Alice'', and ``the card''), rather than relying on pronouns. If the referent would be unclear to a careful human, the intended gold label is \texttt{ABSTAIN} (reason code \texttt{INSUFFICIENT\_CONTEXT}).

\paragraph{Construction principle.}
SECOND\_ORDER families follow the same \(2 \times 2\) structure. For the strict headline subset, we use the XOR pattern in Section~\ref{subsec:xor}. For additional \texttt{ABSTAIN}-labelled families, we require that independent annotators agree that a careful human cannot answer from the photo.

\subsection{Data collection}

All images were collected by the authors in and around the NYU campus. Indoor scenes were photographed in student dormitory spaces (bedrooms, kitchens, laundry rooms, corridors, stairwells, and common areas). Outdoor scenes were photographed in the surrounding city (pavements, street crossings, car parks, and neighbourhood parks). Images were captured at approximately human eye height using a smartphone camera. The release includes per-image metadata and editing provenance.

\paragraph{Ethics and privacy.}
Because images include dormitory interiors and campus environments, we took steps to limit privacy exposure. Faces of bystanders are either absent, turned away, or too small to identify at release resolution. No images contain legible personal identifiers (student IDs, mail labels, or screen contents showing private information). All photographs were taken by the authors with the knowledge of any participants who appear. Evaluation code is released under the MIT license; the dataset (images and metadata) is released under CC~BY~4.0.

\section{Evaluation and scoring}
\label{sec:evaluation}

\subsection{Gold labels and cells}

For the strict XOR headline subset, gold labels are fixed by construction (Section~\ref{subsec:xor}). We refer to the four cells as:
\begin{itemize}[noitemsep, topsep=3pt]
  \item \texttt{BASE}: \((I^0, q^0)\) with gold \texttt{VISIBLY\_FALSE}
  \item \texttt{TEXT\_FLIP}: \((I^0, q^1)\) with gold \texttt{VISIBLY\_TRUE}
  \item \texttt{IMAGE\_FLIP}: \((I^1, q^0)\) with gold \texttt{VISIBLY\_TRUE}
  \item \texttt{DOUBLE\_FLIP}: \((I^1, q^1)\) with gold \texttt{VISIBLY\_FALSE} (diagnostic)
\end{itemize}

Models may output \texttt{ABSTAIN} when they cannot decide with reasonable confidence from the image and question.

\subsection{Confidence-aware accuracy with abstention (CAA)}
\label{subsec:caa}

Let item \(i\) have gold label \(y_i \in \{\texttt{VISIBLY\_TRUE}, \texttt{VISIBLY\_FALSE}\}\). A model outputs \(\hat{y}_i \in \{\texttt{VISIBLY\_TRUE}, \texttt{VISIBLY\_FALSE}, \texttt{ABSTAIN}\}\) and a confidence \(\hat{c}_i \in [0,1]\). (For evaluation, \(\hat{c}_i\) is used only when \(\hat{y}_i \neq \texttt{ABSTAIN}\).)

We define confidence-aware accuracy with abstention (CAA) using a partial credit parameter \(\alpha \in [0,1]\):
\[
\text{score}_i =
\begin{cases}
\alpha & \text{if } \hat{y}_i=\texttt{ABSTAIN},\\
\hat{c}_i & \text{if } \hat{y}_i=y_i,\\
0 & \text{if } \hat{y}_i\neq y_i \text{ and } \hat{y}_i\neq \texttt{ABSTAIN}.
\end{cases}
\]
Then
\[
\text{CAA}=\frac{1}{N}\sum_{i=1}^{N}\text{score}_i.
\]

\paragraph{Justification.}
CAA rewards high confidence when correct, gives zero credit for incorrect answers regardless of confidence, and gives fixed partial credit for abstention. This avoids a gaming vulnerability where low-confidence wrong answers could score well under alternative formulations. The design is inspired by the selective prediction literature, where models trade coverage for reduced risk \citep{elyaniv_selective_2010,geifman_selectivenet_2019}. We use \(\alpha=0.25\) by default to give abstention a small but non-trivial value, reflecting that withholding judgment can be preferable to a guess in safety-relevant settings.

Headline CAA is computed on the three headline cells only (\texttt{BASE}, \texttt{TEXT\_FLIP}, \texttt{IMAGE\_FLIP}).

\paragraph{Worked example.}
Consider one family's three headline cells. Suppose a model returns:
\begin{itemize}[noitemsep, topsep=2pt]
  \item BASE (gold \texttt{VF}): predicts \texttt{VISIBLY\_FALSE} with confidence 0.85 $\to$ score $= 0.85$.
  \item TEXT\_FLIP (gold \texttt{VT}): predicts \texttt{ABSTAIN} $\to$ score $= \alpha = 0.25$.
  \item IMAGE\_FLIP (gold \texttt{VT}): predicts \texttt{VISIBLY\_FALSE} with confidence 0.70 $\to$ score $= 0$ (wrong label).
\end{itemize}
This family contributes $(0.85 + 0.25 + 0) / 3 = 0.367$ to CAA. The abstention on TEXT\_FLIP earns more than the confident wrong answer on IMAGE\_FLIP, illustrating why withholding judgment can be preferable to guessing.

\subsection{Minimal edit flip rates (MEFR)}

We measure sensitivity to minimal edits along each axis, conditioning on correctness of the base cell. For correctness checks, \texttt{ABSTAIN} counts as incorrect.

Let \(c^{ab}_f=\mathbf{1}[\hat{y}^{ab}_f=y^{ab}_f]\) for family \(f\) and cell \((a,b)\).

\[
\text{I\_MEFR}=
\frac{\sum_{f=1}^{F}\mathbf{1}\big[c^{00}_f=1 \land c^{10}_f=1\big]}
{\sum_{f=1}^{F}\mathbf{1}\big[c^{00}_f=1\big]},
\quad
\text{T\_MEFR}=
\frac{\sum_{f=1}^{F}\mathbf{1}\big[c^{00}_f=1 \land c^{01}_f=1\big]}
{\sum_{f=1}^{F}\mathbf{1}\big[c^{00}_f=1\big]}.
\]
\[
\text{MEFR}=\tfrac{1}{2}(\text{I\_MEFR}+\text{T\_MEFR}).
\]

\paragraph{Denominators.}
Because MEFR conditions on correctness of the BASE cell, the effective denominator (the number of families with correct BASE predictions) can vary across models, especially when abstention is frequent. We report MEFR denominators alongside selective prediction diagnostics in Table~\ref{tab:selective_diag}.

\subsection{Confidence-ranked selective prediction score (SelRank)}
\label{subsec:selrank}

We evaluate selective prediction using confidence-ranked answering on headline cells. We consider only \emph{answered} predictions (\texttt{VISIBLY\_TRUE} or \texttt{VISIBLY\_FALSE}) and exclude \texttt{ABSTAIN} from coverage.

Sort answered items by confidence in descending order. For each prefix length \(k\), compute coverage \(\mathrm{cov}(k)=k/n\) and answered accuracy \(\mathrm{acc}(k)\). Let \(A_{\text{model}}\) be the area under the answered accuracy versus coverage curve (trapezoidal rule). Let \(p\) be the overall answered accuracy (the flat baseline).

This is closely related to the standard area under the risk-coverage curve (AURC) used in selective classification \citep{elyaniv_selective_2010,varshney_selective_2022}, since \(\mathrm{risk}(k)=1-\mathrm{acc}(k)\). We use a normalized ``gain'' formulation where higher is better (unlike standard AURC where lower is better), and name it SelRank to avoid confusion.

We report a normalized score (upper-capped at \(1\)):
\[
\mathrm{SelRank}=\min\left(1,\ \frac{A_{\text{model}}-p}{1-p}\right).
\]
This value is \(0\) when confidence ranking is no better than the flat baseline, and increases when higher confidence corresponds to higher correctness. Negative values indicate that confidence ranking is \emph{anti}-informative (lower-confidence answers tend to be more correct).

\subsection{Multi-agent / second-order accuracy (ToMAcc)}

On the MULTI\_AGENT / SECOND\_ORDER strict subset, we report:
\[
\text{ToMAcc}=\frac{\text{number of correct predictions on SECOND\_ORDER items}}{\text{total number of SECOND\_ORDER items}}.
\]
For ToMAcc, \texttt{ABSTAIN} counts as incorrect.

\subsection{Double flip diagnostic accuracy (DFAcc)}

We report diagnostic accuracy on the DOUBLE\_FLIP cell:
\[
\text{DFAcc}=\frac{\text{number of correct predictions on DOUBLE\_FLIP}}{\text{total number of DOUBLE\_FLIP items evaluated}}.
\]
DFAcc is reported separately and is not included in the final composite score.

\subsection{Final composite score}

The final benchmark score combines four headline components with fixed weights:
\[
\textsc{FinalScore}=0.70\cdot \text{CAA}
+0.15\cdot \text{MEFR}
+0.10\cdot \text{SelRank}
+0.05\cdot \text{ToMAcc}.
\]

\paragraph{Weight choice.}
These coefficients reflect a default prioritization for safety-first visibility reasoning: correctness with calibrated abstention is the primary objective (70\%), robustness to minimal edits is next (15\%), confidence ranking for selective answering is third (10\%), and second-order performance is included as a small but distinct capability axis (5\%). The weighting is deliberately conservative---we assign the bulk of the score to the metric most directly tied to deployment safety (CAA). Because different applications may prioritize differently (e.g., weighting MEFR higher for robustness-critical deployments), we report all component metrics separately so that users can reweight for their setting.

If no SECOND\_ORDER items are present in an evaluation run, weights are renormalized over the remaining terms.

\section{Prompt template}
\label{sec:prompt}

We used a single, explicit instruction block (provided as the user message) to standardise how models interpret the task, handle ambiguity, and format outputs. Two parts of the prompt deserve emphasis. First, the referring-expression rules ensure that the target entity can be identified from the image plus the question text alone, without any hidden annotation channel (such as arrows or markings). Second, the instruction ``Do not require arrows, circles, or any on-image markings'' clarifies that the benchmark does not assume any visual markup; questions must be answerable from the unedited photo.

The prompt is shown verbatim below, with \texttt{\{question\}} substituted per item.

\begin{quote}
\begin{lstlisting}
You are given one photo and one question. Decide whether the
question's claim is true using only pixels in this photo.
Do not guess and do not use world knowledge.

Observer

The observer is the camera unless the question explicitly
names a person in the image as the observer.

How to interpret questions

Treat questions like "Is X visible/readable/clear?" or
"Is person A looking at X?" as a true/false claim you must
judge from the image.

For open-ended content questions (eg "what's in the box?",
"what time is it?"), convert them into: "Is that information
visually determinable from this photo right now?"

Referring expressions (who is "he/she/they/it")

Do not require arrows, circles, or any on-image markings.

If the question uses pronouns or vague references, you must
be able to identify the referent unambiguously from the image
plus the question text alone.

Acceptable disambiguation: only one plausible referent
exists, or the question uniquely specifies it via clear
visual descriptors (eg "the man on the left", "the woman
in the red jacket", "the cat on the sofa").

If multiple referents are plausible and the text does not
uniquely pick one, choose ABSTAIN with
reason_code=INSUFFICIENT_CONTEXT.

Output format

Return exactly one JSON object (no prose, no markdown,
no trailing commas) with keys in this order:

{
  "label":       "VISIBLY_TRUE | VISIBLY_FALSE | ABSTAIN",
  "reason_code": "GAZE_DIRECTION | OCCLUSION
                | OUT_OF_FRAME | LIGHTING_DISTANCE
                | INHERENTLY_NONVISUAL
                | AUGMENTED_VISION_REQUIRED
                | INSUFFICIENT_CONTEXT
                | MULTI_AGENT_SECOND_ORDER | NONE",
  "confidence":  <number 0.0 to 1.0>
}

Label meanings

VISIBLY_TRUE:  the claim is clearly true from pixels.
VISIBLY_FALSE: the claim is clearly false from pixels.
ABSTAIN:       you cannot decide true vs false with
               reasonable confidence from this image.

Reason codes

If label="VISIBLY_TRUE", set reason_code="NONE".

If label="ABSTAIN", pick exactly one reason_code explaining
what prevents a decision.

If label="VISIBLY_FALSE":

  If the claim is false because the opposite is clearly true
  (eg the question asserts "not visible" but it is plainly
  visible), you may set reason_code="NONE".

  Otherwise pick exactly one limiting-factor reason_code
  explaining why the claim is false.

Precedence if multiple apply:

  OCCLUSION > OUT_OF_FRAME > GAZE_DIRECTION
  > LIGHTING_DISTANCE > AUGMENTED_VISION_REQUIRED
  > INHERENTLY_NONVISUAL > INSUFFICIENT_CONTEXT
  > MULTI_AGENT_SECOND_ORDER

Transparent clear glass is non-occluding;
frosted/translucent counts as occluding for recognition.

Confidence

Confidence is your probability that your chosen label is
correct (0.0 to 1.0).

Use your own internal thresholding. If you cannot decide
with reasonable confidence, choose ABSTAIN.

Question: {question}
\end{lstlisting}
\end{quote}

\section{Experimental setup}
\label{sec:setup}

We evaluated nine vision-language models on \benchmarkname{}, organized into three groups:
\begin{itemize}[noitemsep, topsep=3pt]
  \item \textbf{Flagship closed-source (3):} Gemini 3.1 Pro (Google), GPT-5 (OpenAI), Claude Opus 4.5 (Anthropic)
  \item \textbf{Prior-generation closed-source (3):} GPT-4o (OpenAI), Gemini 2.5 Pro (Google), Claude 3.7 Sonnet (Anthropic)
  \item \textbf{Open-source 8--12B (3):} Gemma 3 12B \citep{gemma3_2025}, InternVL3-8B \citep{internvl3_2025}, Qwen3-VL-8B \citep{qwen3_vl_2025}
\end{itemize}

All models were queried with the prompt in Section~\ref{sec:prompt}. For each family, we ran the four cells (BASE, TEXT\_FLIP, IMAGE\_FLIP, DOUBLE\_FLIP). Headline metrics used the first three cells only, yielding \(3F=300\) scored headline items per model. We used \(\alpha=0.25\) for CAA. SelRank was computed on answered items only, with \texttt{ABSTAIN} excluded from coverage.

Open-source models (Gemma 3 12B, InternVL3-8B, Qwen3-VL-8B) were run on a single NVIDIA RTX 3090 GPU (24\,GB VRAM) using RunPod cloud infrastructure.

\begin{table}[t]
\centering
\small
\fbox{\parbox{0.93\linewidth}{%
\textbf{How to interpret the metrics} \smallskip \\
\begin{tabular}{@{}l@{\ }p{0.82\linewidth}@{}}
\textbf{CAA} & Rewards correct high-confidence answers, gives zero for wrong answers, and grants fixed partial credit ($\alpha{=}0.25$) for \texttt{ABSTAIN}. \\
\textbf{MEFR} & Did the model flip its answer correctly when a minimal edit changed the evidence? Conditioned on getting BASE right. \\
\textbf{SelRank} & Does confidence sort correct answers above incorrect ones? Measures selective-prediction quality (risk-coverage style). \\
\textbf{ToMAcc} & Accuracy on the MULTI\_AGENT / SECOND\_ORDER slice---can the model reason about what one person can see from another's perspective? \\
\end{tabular}
}}
\caption*{} % suppress numbering
\end{table}

\section{Results}
\label{sec:results}

\subsection{Overall performance}

Table~\ref{tab:main_results} reports the main results on the strict headline subset for all nine models. GPT-4o and Gemini 3.1 Pro effectively tie for the highest overall \textsc{FinalScore} (0.728 and 0.727 respectively), followed by Gemini 2.5 Pro (0.678) and GPT-5 (0.625). Claude Opus 4.5 (0.570) forms a middle tier among flagship systems. Among prior-generation models, GPT-4o (0.728) substantially outperforms its tier peers Gemini 2.5 Pro (0.678) and Claude 3.7 Sonnet (0.476). The best open-source model, Gemma 3 12B (0.505), surpasses Claude 3.7 Sonnet, demonstrating that current open-source models at the 8--12B scale can exceed prior-generation closed-source systems on visibility reasoning. InternVL3-8B (0.445) and Qwen3-VL-8B (0.419) occupy the lower range.

Abstention behavior varies substantially across models. GPT-5 abstains most frequently (78 of 300 headline items), followed by Qwen3-VL-8B (50), while Gemini 3.1 Pro abstains least (14 of 300). Qwen3-VL-8B also produced 66 unparsable outputs across all cells (49 in headline cells), reducing its effective coverage. All model outputs were parsed with strict JSON extraction only: we strip leading/trailing whitespace and code fences, then attempt a single \texttt{json.loads} call with no further repair or heuristic extraction, so unparsable outputs reflect genuine format-following failures rather than aggressive filtering. Because CAA awards partial credit for abstention, we report abstention counts alongside all headline metrics.

\begin{table}[t]
\centering
\small
\setlength{\tabcolsep}{3.5pt}
\caption{Headline results on \benchmarkname{} (300 scored headline items per model). CAA uses \(\alpha=0.25\). SelRank is a normalized selective ranking score on answered items (negative values indicate anti-informative confidence ranking). \texttt{ABS} is the number of abstentions among the 300 headline items. Models are grouped into three tiers (3/3/3): flagship closed-source, prior-generation closed-source, and open-source 8--12B.}
\label{tab:main_results}
\begin{tabular}{l
S[table-format=1.3]
S[table-format=1.3]
S[table-format=1.3]
S[table-format=1.3]
S[table-format=1.3]
S[table-format=2.3]
S[table-format=1.3]
S[table-format=2.0]
}
\toprule
\textbf{Model} & {\textbf{Final}} & {\textbf{CAA}} & {\textbf{I\_MEFR}} & {\textbf{T\_MEFR}} & {\textbf{MEFR}} & {\textbf{SelRank}} & {\textbf{ToMAcc}} & {\textbf{ABS}} \\
\midrule
Gemini 3.1 Pro    & 0.727 & 0.760 & 0.659 & 0.871 & 0.765 &  0.394 & 0.810 & 14 \\
GPT-5             & 0.625 & 0.622 & 0.793 & 0.845 & 0.819 &  0.237 & 0.857 & 78 \\
Claude Opus 4.5   & 0.570 & 0.580 & 0.570 & 0.671 & 0.620 &  0.374 & 0.667 & 18 \\
\midrule
GPT-4o            & 0.728 & 0.769 & 0.800 & 0.893 & 0.847 &  0.144 & 0.952 & 16 \\
Gemini 2.5 Pro    & 0.678 & 0.747 & 0.743 & 0.892 & 0.818 & -0.106 & 0.857 & 18 \\
Claude 3.7 Sonnet & 0.476 & 0.508 & 0.507 & 0.493 & 0.500 &  0.192 & 0.524 & 30 \\
\midrule
Gemma 3 12B       & 0.505 & 0.543 & 0.424 & 0.644 & 0.534 &  0.087 & 0.714 & 25 \\
InternVL3-8B      & 0.445 & 0.498 & 0.610 & 0.373 & 0.492 &  0.018 & 0.429 & 24 \\
Qwen3-VL-8B      & 0.419 & 0.509 & 0.307 & 0.180 & 0.243 &  0.033 & 0.450 & 50 \\
\bottomrule
\end{tabular}
\end{table}

\subsection{Minimal edit sensitivity}

T\_MEFR exceeds I\_MEFR for six of nine models, indicating that text-flipped questions are generally handled more reliably than minimal image edits. This asymmetry is most pronounced for GPT-4o (T\_MEFR 0.893 vs.\ I\_MEFR 0.800) and Gemini 2.5 Pro (0.892 vs.\ 0.743). Three models reverse the pattern: InternVL3-8B (I\_MEFR 0.610 vs.\ T\_MEFR 0.373), Qwen3-VL-8B (0.307 vs.\ 0.180), and Claude 3.7 Sonnet (0.507 vs.\ 0.493), though the last reversal is marginal.

GPT-4o achieves the highest overall MEFR (0.847), followed by GPT-5 (0.819) and Gemini 2.5 Pro (0.818), indicating robust handling of both edit types. Qwen3-VL-8B shows the lowest MEFR (0.243), partly due to its low T\_MEFR (0.180), suggesting difficulty with negated questions.

\subsection{Abstention, coverage, and confidence ranking}

Since SelRank is answered-only, it is useful to also report answered coverage and answered accuracy. Table~\ref{tab:selective_diag} reports answered fraction, answered-only accuracy, the MEFR denominator (families where BASE is correct), and the raw normalized SelRank.

\begin{table}[t]
\centering
\small
\setlength{\tabcolsep}{5pt}
\caption{Selective prediction and MEFR diagnostics on headline items. \textbf{Answered} counts non-\texttt{ABSTAIN} outputs among 300 headline items (for Qwen3-VL-8B, 49 headline items were unparsable and excluded). \textbf{Coverage} is answered/300. \textbf{AnsAcc} is accuracy on answered items only. \textbf{SelRank$_{\text{raw}}$} is the normalized selective ranking score before capping. \textbf{MEFR d} is the number of families with correct BASE prediction. Model grouping follows Table~\ref{tab:main_results}.}
\label{tab:selective_diag}
\begin{tabular}{l
S[table-format=3.0]
S[table-format=2.1]
S[table-format=1.3]
S[table-format=2.3]
S[table-format=2.0]
}
\toprule
\textbf{Model} & {\textbf{Answered}} & {\textbf{Cov (\%)}} & {\textbf{AnsAcc}} & {\textbf{SelRank$_{\text{raw}}$}} & {\textbf{MEFR d}} \\
\midrule
Gemini 3.1 Pro    & 286 & 95.3 & 0.804 &  0.394 & 85 \\
GPT-5             & 222 & 74.0 & 0.851 &  0.237 & 58 \\
Claude Opus 4.5   & 282 & 94.0 & 0.684 &  0.374 & 79 \\
\midrule
GPT-4o            & 284 & 94.7 & 0.831 &  0.144 & 75 \\
Gemini 2.5 Pro    & 282 & 94.0 & 0.794 & -0.106 & 74 \\
Claude 3.7 Sonnet & 270 & 90.0 & 0.589 &  0.192 & 73 \\
\midrule
Gemma 3 12B       & 275 & 91.7 & 0.618 &  0.087 & 59 \\
InternVL3-8B      & 273 & 91.0 & 0.553 &  0.018 & 59 \\
Qwen3-VL-8B      & 201 & 67.0 & 0.582 &  0.033 & 72 \\
\bottomrule
\end{tabular}
\end{table}

Several patterns emerge. First, GPT-5 achieves the highest answered-only accuracy (0.851) but with unusually low coverage (74.0\%, 78 abstentions), indicating a cautious strategy that substantially reduces its composite score. Second, GPT-4o achieves the second-highest answered accuracy (0.831) with high coverage (94.7\%) and moderate calibration (SelRank 0.144). Third, Gemini 3.1 Pro achieves the best confidence calibration (SelRank 0.394) alongside strong answered accuracy (0.804, 95.3\% coverage), meaning higher-confidence answers tend to be more accurate. By contrast, Gemini 2.5 Pro achieves similar answered accuracy (0.794) and coverage (94.0\%) but has anti-informative confidence ranking (SelRank $-0.106$), meaning its lower-confidence answers tend to be more correct. Among open-source models, Qwen3-VL-8B shows the lowest effective coverage (67.0\%) due to 49 unparsable headline outputs, though its answered-only accuracy (0.582) exceeds that of InternVL3-8B (0.553).

\subsection{Multi-agent and second-order subset}

ToMAcc is computed on the MULTI\_AGENT / SECOND\_ORDER slice (7 families, 21 headline items). GPT-4o achieves the best performance on this subset (0.952), followed by GPT-5 and Gemini 2.5 Pro (both 0.857) and Gemini 3.1 Pro (0.810). Among open-source models, Gemma 3 12B scores 0.714, while Claude 3.7 Sonnet (0.524), Qwen3-VL-8B (0.450), and InternVL3-8B (0.429) perform near chance level. Claude Opus 4.5 scores 0.667. The strong performance of GPT-4o on second-order reasoning is notable, with all three prior-generation and flagship Google and OpenAI models substantially outperforming open-source alternatives.

\paragraph{Confidence intervals.} With only 21 items, ToMAcc estimates have wide confidence intervals. Using Wilson's method, the 95\% CI for ToMAcc = 0.952 (20/21) is approximately [0.77, 0.99], and for ToMAcc = 0.429 (9/21) is approximately [0.24, 0.64]. Rankings within this subset should be interpreted cautiously; expanding the SECOND\_ORDER slice is a priority for future dataset releases.

\subsection{Diagnostic: double flip}

We also record DOUBLE\_FLIP behavior \((I^1,q^1)\) for diagnosis, but it is not included in \textsc{FinalScore}. The DOUBLE\_FLIP cell can identify families where composing edits yields unexpected interactions, or where an intended atomic edit incidentally changes additional visibility factors. In this release we treat DOUBLE\_FLIP primarily as a diagnostic signal and leave systematic family auditing based on DOUBLE\_FLIP to future work.

\subsection{Open-source versus closed-source}

A substantial performance gap exists between flagship closed-source models and open-source models at the 8--12B scale. GPT-4o (0.728) and Gemini 3.1 Pro (0.727) outperform the best open-source model, Gemma 3 12B (0.505), by over 0.22 in composite score (roughly 30\% relative). The gap is particularly pronounced on second-order reasoning (ToMAcc), where GPT-4o (0.952) and Gemini 3.1 Pro (0.810) far exceed all open-source alternatives (Gemma 3 12B: 0.714, InternVL3-8B: 0.429, Qwen3-VL-8B: 0.450).

However, the open-source tier shows meaningful differentiation. Gemma 3 12B (0.505) surpasses one prior-generation closed-source system---Claude 3.7 Sonnet (0.476)---demonstrating that visibility reasoning is not exclusively a closed-source capability. At the other extreme, Qwen3-VL-8B shows the lowest MEFR (0.243) and produced 66 unparsable outputs across all cells, suggesting output-format compliance issues that limit effective evaluation. InternVL3-8B (0.445) falls between its open-source peers, with notably high image-flip robustness (I\_MEFR 0.610) but weak text-flip handling (T\_MEFR 0.373).

\section{Discussion}

The evaluation of nine models on \benchmarkname{} reveals three findings about the state of visibility reasoning in current vision-language models.

\paragraph{Flagship versus open-source gap.} The top closed-source models (GPT-4o, Gemini 3.1 Pro) outperform the best open-source model (Gemma 3 12B) by roughly 30\% in relative composite score (0.728 vs.\ 0.505). The gap is largest on second-order reasoning, where GPT-4o (ToMAcc 0.952) and Gemini 3.1 Pro (0.810) substantially exceed open-source alternatives. However, Gemma 3 12B surpasses one prior-generation closed-source system, suggesting that the gap is narrowing at the 8--12B parameter scale.

\paragraph{Text-flip versus image-flip asymmetry.} T\_MEFR exceeds I\_MEFR for six of nine models, suggesting that models are generally better at tracking logical negation in text than at detecting subtle visual changes between minimally edited photos. This asymmetry has practical implications: text augmentation (rephrasing questions, testing negations) may be a more effective robustness intervention than visual augmentation for current systems. Three models reverse the pattern---InternVL3-8B, Qwen3-VL-8B, and Claude 3.7 Sonnet---though the last reversal is marginal (0.507 vs.\ 0.493). Qwen3-VL-8B shows the lowest T\_MEFR (0.180), suggesting a floor effect in which some models cannot reliably handle negated questions.

\paragraph{Calibration variability.} GPT-4o and Gemini 2.5 Pro achieve similar accuracy (CAA 0.769 vs.\ 0.747) yet differ sharply in selective prediction quality (SelRank 0.144 vs.\ $-0.106$). Gemini 3.1 Pro achieves the best calibration overall (SelRank 0.394) alongside strong accuracy (CAA 0.760). This dissociation implies that accuracy alone is insufficient for deployment decisions: a system whose confidence scores do not rank correctness cannot safely defer uncertain predictions. The connection to selective prediction literature \citep{elyaniv_selective_2010,geifman_selectivenet_2019} is direct---SelRank measures exactly the property needed for risk-coverage trade-offs in practice.

\paragraph{Limitations.}
The current release has several limitations. First, 100 families provide limited statistical power for fine-grained comparisons, particularly on the 21-item SECOND\_ORDER slice. Second, all images were collected in a narrow set of environments (campus buildings and nearby streets), so results may not generalize to other settings such as rural, industrial, or indoor-medical scenes. Third, the composite score uses fixed weights; we have not yet conducted a sensitivity analysis over alternative weightings. Fourth, the benchmark uses short yes/no questions to tightly control claims, which improves interpretability but does not cover longer multi-step reasoning. Fifth, Qwen3-VL-8B produced 66 unparsable outputs, reducing its effective evaluation set and potentially underestimating its capabilities. More broadly, low scores on \benchmarkname{} can reflect either weak visibility reasoning (a capability limitation) or poor structured-output compliance (a format-following limitation); the two are conflated in any benchmark that requires JSON. We mitigate this by reporting unparsable counts explicitly and using only minimal parsing (whitespace/code-fence stripping), but future work could explore constrained decoding or function-calling APIs to isolate capability from formatting.

\paragraph{Open problems.}
The three findings above point directly to open research questions:
\begin{enumerate}[noitemsep, topsep=3pt]
  \item \textbf{Closing the image-flip gap.} Why are minimal visual edits harder to detect than text negations, and can training on controlled image perturbations improve I\_MEFR without sacrificing T\_MEFR?
  \item \textbf{Confidence ranking that matches accuracy.} GPT-4o and Gemini 2.5 Pro achieve similar accuracy yet diverge sharply on SelRank. What training or calibration interventions produce confidence scores that reliably sort correct above incorrect answers?
  \item \textbf{Scalable second-order evaluation.} With only 21 SECOND\_ORDER items, current rankings have wide confidence intervals. Building larger, more diverse multi-agent slices---potentially with synthetic scene generation---is needed to draw robust conclusions about perspective reasoning.
\end{enumerate}

\section{Conclusion}

We introduced \benchmarkname{}, a benchmark for visibility and perspective reasoning in images with a controlled \(2 \times 2\) family construction, explicit abstention, and a dedicated MULTI\_AGENT / SECOND\_ORDER slice. We evaluate nine vision-language models across 100 families. GPT-4o and Gemini 3.1 Pro achieve the highest composite scores, while the best open-source model (Gemma 3 12B) surpasses one prior-generation closed-source system, demonstrating that visibility reasoning capabilities are beginning to transfer to the open-source tier at the 8--12B scale. Future work will expand the benchmark with additional families, a larger SECOND\_ORDER slice, and images from broader environments to strengthen statistical power and generalizability.

\section*{Reproducibility checklist}

\begin{itemize}[noitemsep, topsep=3pt]
  \item \textbf{Models:} Gemini 3.1 Pro, GPT-5, Claude Opus 4.5, GPT-4o, Gemini 2.5 Pro, Claude 3.7 Sonnet, Gemma 3 12B, InternVL3-8B, Qwen3-VL-8B.
  \item \textbf{API call dates:} All closed-source model runs were completed between January and March 2026. Exact dates and model version strings (API model IDs) are recorded in the released run metadata.
  \item \textbf{Temperature:} Default API temperature for all models (no manual override).
  \item \textbf{Image resolution:} Images were sent at their original resolution (3024$\times$4032) to closed-source APIs. Open-source models applied their default internal resizing.
  \item \textbf{Retry policy:} Failed API calls (timeouts, rate limits) were retried up to 3 times with exponential backoff. No retries were made based on output content.
  \item \textbf{Open-source hardware:} Single NVIDIA RTX 3090 (24\,GB VRAM) via RunPod.
  \item \textbf{Prompt:} Verbatim in Section~\ref{sec:prompt}; identical across all models.
  \item \textbf{Code and data:} Released at \url{https://github.com/neilt93/Paper-with-Davis}.
\end{itemize}

\section*{Acknowledgements}
I thank Prof.\ Ernest Davis of New York University (davise@cs.nyu.edu) for detailed feedback on benchmark design, paper presentation, and evaluation methodology.

\bibliographystyle{plainnat}
\bibliography{references}

\end{document}